\documentclass[lettersize,journal]{IEEEtran}
\usepackage{amsmath,amsfonts}
\usepackage{algorithmic}
\usepackage{algorithm}
\usepackage{array}
\usepackage[caption=false,font=normalsize,labelfont=sf,textfont=sf]{subfig}
\usepackage{textcomp}
\usepackage{stfloats}
\usepackage{url}
\usepackage{verbatim}
\usepackage{graphicx}
\usepackage{cite}
\usepackage{booktabs}
\usepackage{amsthm,amsmath,amssymb}
\usepackage{multirow}
\usepackage{color, soul, framed}

\usepackage{color}

\usepackage{makecell}
\usepackage{mathrsfs}
\usepackage{pifont}

\usepackage{threeparttable}
\usepackage{bm}
\usepackage{caption}
\usepackage{lipsum}
\usepackage{ragged2e}
\usepackage{colortbl}
\definecolor{wjbest}{rgb}{0.96, 0.57, 0.58}
\definecolor{wjsecond}{rgb}{0.98, 0.78, 0.57}
\definecolor{wjthird}{rgb}{1.0, 1.0, 0.56}

\definecolor{gscolor}{rgb}{1.0,0.6,0.0} %

\usepackage{xcolor, soul}
\sethlcolor{green}
\usepackage{makecell}

\begin{document}

\title{GaussianHead: High-fidelity Head Avatars with Learnable Gaussian Derivation}

\author{Jie Wang, Jiu-Cheng Xie, Xianyan Li, Feng Xu, Chi-Man Pun,   and Hao Gao% <-this % stops a space
\thanks{Jie Wang, Jiu-Cheng Xie, Xianyan Li and Hao Gao are with the School of Automation and the School of Artificial Intelligence, Nanjing University of Posts and Telecommunications, Nanjing, 210023, China.
E-mail: chieh.wangs@gmail.com, jiuchengxie@gmail.com, 974598lxy@gmail.com, tsgaohao@gmail.com.}% <-this % stops a space
\thanks{Feng Xu is with the School of Software and BNRist, Tsinghua University, Beijing 100084, China.
E-mail: xufeng2003@gmail.com.}% <-this % stops a space
\thanks{Chi-Man Pun is with the Department of Computer and Information Science, University of Macau, Taipa, Macau.  
E-mail: cmpun@um.edu.mo.}% <-this % stops a space
\thanks{Jiu-Cheng Xie and Hao Gao are the corresponding authors.}}

\markboth{Journal of \LaTeX\ Class Files,~Vol.~14, No.~8, August~2021}%
{Shell \MakeLowercase{\textit{et al.}}: A Sample Article Using IEEEtran.cls for IEEE Journals}


\let\oldtwocolumn\twocolumn
\renewcommand\twocolumn[1][]{%
    \oldtwocolumn[{#1}{
    \begin{center}
           \includegraphics[width=1\textwidth]{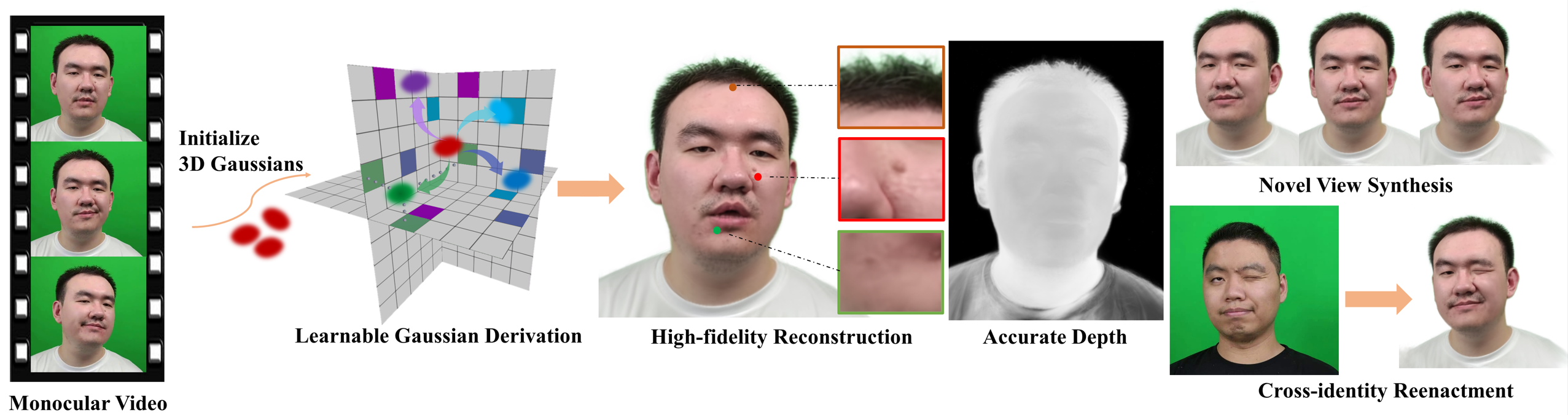}
           \captionof{figure}{Based on anisotropic 3D Gaussians and the learnable derivation strategies, our method learns an identity-specific head avatar from a monocular video of the corresponding subject. The proposed GaussianHead demonstrates outstanding performance in self-reconstruction, cross-identity reenactment, and novel-view synthesis tasks.}
           \label{fig:teaser}
        \end{center}
    }]
}

\maketitle

% \twocolumn[{%

% \renewcommand\twocolumn[1][]{#1}%
% \maketitle
% \begin{center}
%  \centering
%  \includegraphics[scale=0.2]{image/teaser.pdf}
%  \captionof{figure}{Based on anisotropic 3D Gaussians and the learnable derivation strategies, our method learns an identity-specific head avatar from a monocular video of the corresponding subject. The proposed GaussianHead demonstrates outstanding performance in self-reconstruction, cross-identity reenactment, and novel-view synthesis tasks.}
% \end{center}
% % \maketitle
% }]

\begin{abstract}
Creating lifelike 3D head avatars and generating compelling animations for diverse subjects remain challenging in computer vision. This paper presents GaussianHead, which models the active head based on anisotropic 3D Gaussians. Our method integrates a motion deformation field and a single-resolution tri-plane to capture the head's intricate dynamics and detailed texture. Notably, we introduce a customized derivation scheme for each 3D Gaussian, facilitating the generation of multiple ``doppelgangers'' through learnable parameters for precise position transformation. This approach enables efficient representation of diverse Gaussian attributes and ensures their precision. Additionally, we propose an inherited derivation strategy for newly added Gaussians to expedite training. Extensive experiments demonstrate GaussianHead's efficacy, achieving high-fidelity visual results with a remarkably compact model size ($\approx 12$ MB). Our method outperforms state-of-the-art alternatives in tasks such as reconstruction, cross-identity reenactment, and novel view synthesis. The source code is available at: \url{https://github.com/chiehwangs/gaussian-head}.
\end{abstract}


\section{Introduction}\label{sec:introduction}
\IEEEPARstart{C}{REATING} personalized head avatars is very useful in wide-ranging applications like virtual reality, remote meetings, movies, and games. Previous works usually rely on synchronized multi-view images of heads to reconstruct photorealistic 3D avatars \cite{nersemble, hq3d, binocular}. 
The harsh capturing requirement hinders the wide range of popularity of these techniques among the public. 
An emergent exploration in this field focuses on using monocular videos to build personalized head avatars, which is much more convenient. Since almost all daily used smartphones now integrate high-quality cameras, this kind of technique has the potential to be used by all end users and may establish more applications like social communication and short video sharing. 
Nevertheless, the variable geometry and complex appearance of human heads make it very challenging to achieve precise modeling from single-view captures.    

Early methods in this direction predominantly rely on pre-trained parametric models \cite{3dmm1999_correct,bfm,flame} to create coarse facial geometry \cite{real-facial-capture} and corresponding appearance \cite{real-facial-capture-2}. While limited in fine details, these methods lay a solid foundation for accurate facial motion capture. Building implicit head avatars using Signed Distance Fields (SDF) \cite{idr,volsdf} helps overcome the deficiencies of parametric models and achieve better head geometry \cite{imavatar,avatarSDF}. Nevertheless, implicit surfaces lack enough expressive power for fine structures and suffer from inefficient rendering. To this end, combining structured data containers and volume rendering to build volumetric neural radiance fields gradually becomes the current mainstream practice. When it comes to the scope of head construction, Several methods \cite{nerfblendshape, avatarmav} utilize multi-resolution neural voxels as facial expression bases, combining them with expression parameters to produce visually compelling dynamic head avatars. Other approaches \cite{er_nerf, synctalk, gaussiantalker} leverage the tri-plane for feature storage, which produces impressive rendering effects and further reduces the quantity of model parameters \cite{k-planes, hexplane}.

% 添加了参考文献 gaussianTalker

% Nevertheless, these methods are constrained by the heavy training overhead associated with neural radiance fields since they require extensive point sampling along rays. 

% In addition, the ``feature dilution" drawbacks resulting from axis-aligned mappings in structured data containers cause unexpected problems to them.

% Nevertheless, these methods are constrained by the heavy training overhead associated with neural radiance fields since they require extensive point sampling along rays.

\begin{figure}[t]
  \centering
   \includegraphics[scale=0.12]{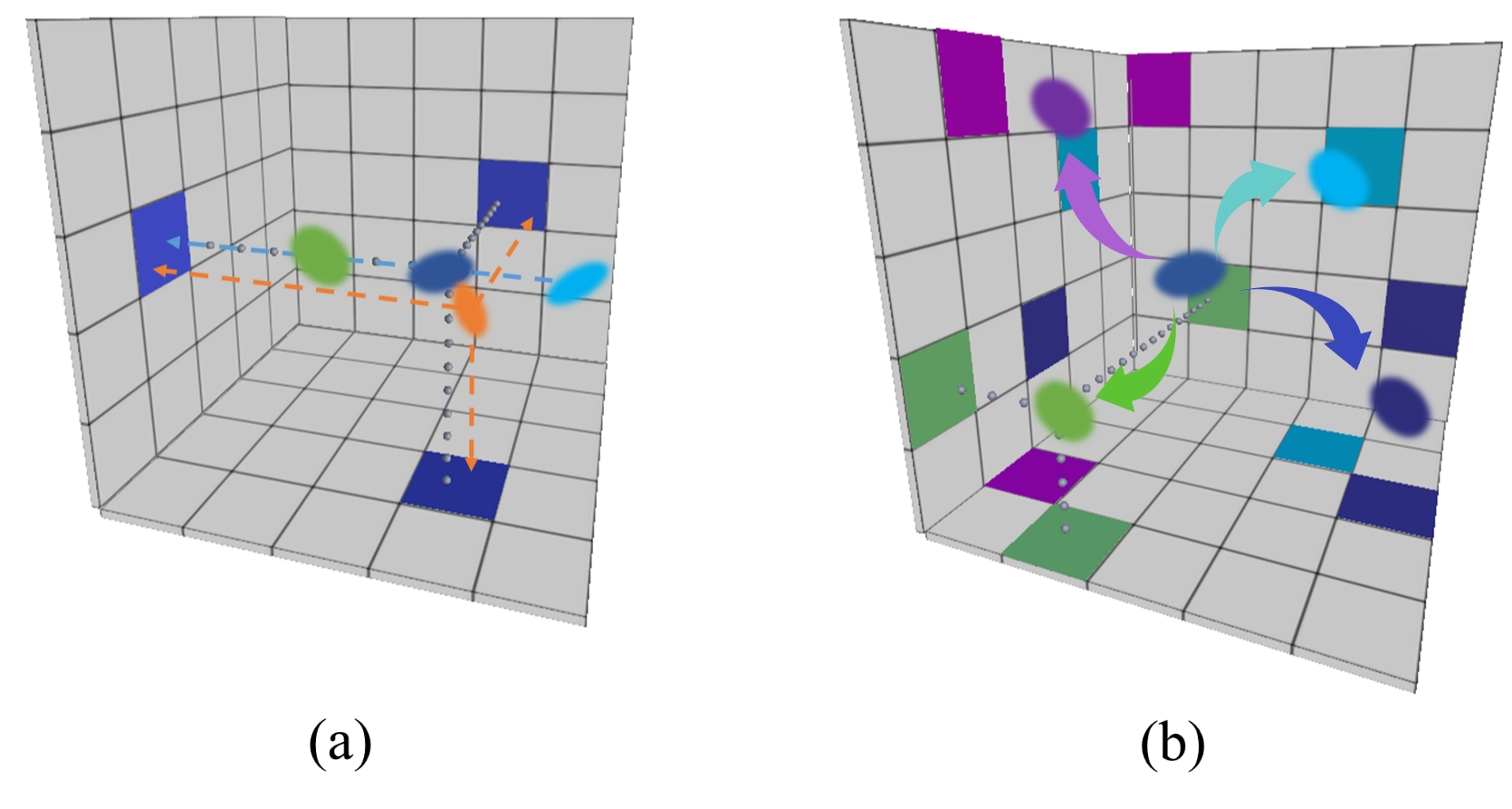}
   \caption{The adoption of axis-aligned mapping in the tri-plane-formed feature representation is accompanied by a severe problem of ``feature dilution'', a visual illustration of that is given in (a). We refer the readers to the third paragraph of the Introduction part for detailed explanations. In contrast, (b) shows our derivation strategy for addressing this problem. Taking the blue Gaussian primitive for example, we first obtain its multiple derivatives through a specific set of learnable transformations and then do feature projection and aggregation. Other 3D Gaussians also undergo the same process.}
   \label{fig:explain-dilution}
\end{figure}

Despite the advantages of tri-plane-formed feature containers, their reliance on a multi-resolution mechanism leaves room for further optimization of the model size. Moreover, the widely adopted axis-aligned mapping is unsuitable in at least two scenarios listed follow. The first case is when multiple primitives are close (the orange and blue blocks in Fig. \ref{fig:explain-dilution}), their projections on each feature plane will be also near. For ease of computation and data saving, each plane is regarded as a grid composed of cells in practice. If these projected positions fall within a single cell, the feature from that single cell will be used to represent information about different primitives, which should be avoided in the pursuit of distinctive feature representation. For the traditional tri-plane, this problem is alleviated by using several tri-planes at different resolutions. The essence of this solution is to reduce the likelihood of overlapped projections by considering different sizes for cells on the feature plane. As a side effect, this solution needs to afford a large number of parameters, which will be proved in Table \ref{tab:multi-res}. The second case is when two or more primitives are in a line that is perpendicular to a certain feature plane (green, blue, and cyan blocks in Fig. \ref{fig:explain-dilution}). Then their axis-aligned projections on the feature plane are always identical, which cannot be simply solved by enlarging or reducing the cell's size. Note that two concurrent works dedicated to the static full-head generation, namely PanoHead\cite{Panohead} and SphereHead\cite{Spherehead}, also report abnormal results caused by this issue when integrating the tri-plane structure in their frameworks. Specifically, they observe the back of the head usually shows the components of the front face (e.g., the eyes appear on the back), the reason for which is points belonging to the front and back of the head are projected to the same location on a certain feature plane, thus causing representation ambiguity. The problem caused by the above two cases is collectively called ``feature dilution'' of the tri-plane in our paper.

% Firstly, primitives on the same projection line (represented by the cyan, blue, and green blocks in Fig. \ref{fig:explain-dilution}) are repetitively mapped to the same position on the feature plane. \textcolor{blue}{Two concurrent works dedicated to the static full-head generation, namely PanoHead\cite{Panohead} and SphereHead\cite{Spherehead}, also report abnormal results caused by this issue when integrating the tri-plane structure in their frameworks. Specifically, they observe the back of the head usually shows the components of the front face (e.g., the eyes appear on the back), the reason for which is points belonging to the front and back of the head are projected to the same location on a certain feature plane, thus causing representation ambiguity.} Secondly, due to the finite resolution of the plane, primitives in close neighboring regions (the orange and red blocks in Fig. \ref{fig:explain-dilution}) may also be projected on the same cell. Consequently, the features stored at a particular cell no longer exclusively or partially correspond to a specific primitive, which reduces the distinguishability among primitives, eventually weakening the representational capacity for sophisticated structures. This limitation is collectively referred to as ``feature dilution" in this paper. 

Feature dilution becomes more pronounced in particular head areas with complex geometry, such as regions obscured by lips, hair, and wrinkles, where denser primitives often exist, accompanied by a significant occurrence of overlapping. TILTED \cite{tilted} aims to achieve a more accurate feature representation of these primitives interpolated on the feature planes by rotating the entire scene simultaneously. While it appears to be effective in static scenarios, shortcomings arise in dynamic scenes. The inherent dynamism of objects disrupts those attempts to address biases in axis-aligned signals by relying on unified scene rotations. Additionally, since rotating the entire object does not change the relative topological structure, it is less effective in mitigating specific feature dilution within overlapped structures.

In this paper, we introduce the GaussianHead, a new method based on deformable 3D Gaussians for faithful head avatar construction and animation. 
It relies on shape-flexible 3D Gaussians and a compact tri-plane feature container, while a novel Gaussian derivation mechanism allows it to achieve high-quality rendering and competitive model size without relying on a multi-resolution paradigm. 
Notably, we disentangle the modeling of head geometry and texture attributes. 
To be concrete, a motion deformation field is first built to fit the whole head shape as well as the dynamic facial movements. In practice, it takes pre-retrieved facial expression coefficients as conditions to transform randomly initialized 3D Gaussians into a posed space. 
Then we employ a compact data container—parameterized single-resolution tri-plane—to store the appearance information of those Gaussians. To resolve the aforementioned feature dilution problem, we design a novel Gaussian derivation strategy. In short, we generate multiple derivations of each core Gaussian in the posed space using the sets of learnable rotation transformations. By acquiring sub-features from these derived doppelgangers and integrating them, we are able to obtain the precise features of the current Gaussians. In contrast to the multi-resolution regime adopted by the original tri-plane, the Gaussian derivation strategy also reduces the likelihood of feature ambiguity but more effectively and economically.
Moreover, a mechanism of inheritance derivation initialization is adopted for newly added Gaussians in the later training phase, improving the convergence speed.

Contributions of this paper are summarized as follows:
\begin{itemize}

\item Leveraging anisotropic 3D Gaussian primitives and a parameterized tri-plane with a single resolution to represent the dynamic head avatar compactly.

\item The novel strategy of deriving Gaussians effectively alleviates the feature dilution caused by axis-aligned mapping, resulting in a more accurate representation of the texture of complex facial structures.

\end{itemize}

\section{Related Work}

% \noindent\textbf{Parametric Head Model.} The 3D morphable model (3DMM) can provide prior knowledge of head structure \cite{INSTA}, reducing uncertainty during neural network learning. Previous parameterized head skinning models decompose the head pose into identity, expression, and pose information and are constructed from a large set of scanned data, capturing low-dimensional facial subspaces using PCA to discover the relationships between blend shape bases and coefficients. FLAME \cite{flame} decomposes the head into a combination of shape, pose, and expression using Linear Blend Skinning (LBS). \cite{3dmm_2,3dmm_3} Compensating for coarse geometric shapes is achieved by adjusting non-rigid mesh deformations and modifying the blend shape bases. With the advancement of deep learning techniques,  \cite{deca,emoca} employ neural networks to predict FLAME model parameters from input images. \cite{3dmm} proposes a semantic disentangled variational autoencoder (SDVAE), further addressing the problem of 3DMM's inability to express skin wrinkles. However, the 3DMMs introduced above are all part of skinning models and cannot fully express the numerous non-rigid structures and delicate textures of the head.

\subsection{Scene Primitives in Reconstruction} 
Whether it is a real-world scenario, the human body, or a head, the foundation of their construction lies in a simple form of scene primitives. Some past approaches rely on implicit primitives. For instance, methods \cite{bakedsdf,idr,deepsdf} use the signed distance function, which builds objects by tracing points located on the zero-level set of the function in space. Algorithms \cite{occupancy,imavatar,occupancy2} leverage the occupancy function to represent 3D surfaces as the continuous decision boundaries of deep neural network classifiers. However, these implicit primitives cannot represent complex head avatars. Methods based on neural radiance fields \cite{nerf,nerface,nerf1, controllable_nerf,nerfcap,nerfplayer} store scenes using network weights. However, the necessary manner of constructing a large number of sample points in each ray always leads to significant training overhead. Explicit scene primitives, such as points \cite{pointavatar,pointnerf,point-recon}, can capture sufficiently complex structures. However, due to the fixed shape of points, detailing can only be achieved by continuously refining point radii and increasing point counts during the training process, which introduces significant storage and training cost. Here, we use 3D anisotropic Gaussians \cite{3d_gaussian} as scene primitives. They are deformable in geometric structure, allowing for the representation of intricate details by adjusting their shapes instead of blindly increasing quantity or reducing radii.

\subsection{Sources for Head Portrait Synthesis} 
The widespread application of head avatars calls for convenient capture modes. some studies \cite{binocular,nersemble,hq3d,gaussianhead_avatar,gaussianavatars} utilize dense multi-view or binocular vision to record dynamic head avatars in space. Specifically, GaussianHeadAvatar \cite{gaussianhead_avatar} and GaussianAvatars \cite{gaussianavatars} use 3D Gaussians combined with multi-view video to construct finely detailed head avatars. Although this brings better multi-view consistency, the difficulty of popularizing capture devices hinders the rapid adoption of such techniques. Meanwhile, NeRSemble \cite{nersemble} reconstructs head dynamics along the temporal axis, sacrificing cross-subject generalization performance. Some portrait synthesis algorithms start with a single image or a set of images, constructing identity-consistent head avatars through image-to-image transformations. Methods \cite{npf-2,nofa,npf,gan_work_1} leverage encoders to encode consistent identities from image sets and regulate outputs by fitting expressions or landmarks from the 3D morphable model. While achieving detailed rendered images, they lack multi-view consistency. Some head avatar methods reconstruct the head by decoupling appearance features\cite{headnerf} or using audio to animate the head avatar\cite{adnerf, radnerf,er_nerf,learn2talk}. Other methods track expressions and poses from monocular videos as reconstruction or reenactment conditions {\cite{face2face}, 
{\cite{imavatar,pointavatar,INSTA,nerfblendshape,SplattingAvatar}}, achieving increasingly realistic results. Similar to them, our approach also constructs the head avatars based on the monocular video of the target subject. 

\subsection{Hybrid Neural Field}
\label{sec:hybrid-net} 
Multiple implicit methods \cite{nerf,imavatar,deepsdf,occupancy} have demonstrated that, as the scene's complexity increases, the learning difficulty for network models also increases. Recently, hybrid neural radiance fields have become popular, which use structured data containers (i.e., tri-plane \cite{eg3d}, hexplane \cite{hexplane,k-planes}, and voxel \cite{instant-ngp}) to store and a decoder to parse the information of scenes. In this manner, the encoding and decoding tasks are undertaken by multiple networks rather than one, reducing the modeling difficulty for each component. However, the commonly employed axis-aligned mappings by these methods will lead to imprecise representation of features. In addition, their multi-scale design has too many parameters, resulting in huge model sizes. We utilize the tri-plane structure and a decoder to represent a head based on Gaussian primitives in 3D space. The highlight of our approach is the proposed Gaussian derivation strategy and relevant adaptive designs to address the limitations of axis-aligned mapping through dispersing sub-features of each primitive, achieving excellent rendering performance with a small model size. We notice a concurrent work \cite{triplane_gaussian} also combines tri-plane with Gaussian splatting but is distinct from us in several aspects: they still query the feature of each Gaussian by projecting it to factored planes along three axes, and they focus on synthesis in static scenarios while ours cares about the dynamic setting.

% For these hybrid neural fields, they mostly adopt a multi-resolution mechanism to achieve excellent visual effects. However, the multi-scale design brings a significant number of parameters, resulting in excessively large model sizes. In addition, the commonly employed axis-aligned mappings can lead to an imprecise representation of features. Our novel Gaussian derivation strategy addresses the limitations of axis-aligned mapping by dispersing sub-features of each primitive, achieving excellent rendering performance even at fine scales without relying on a multi-resolution mechanism.

% However, the multi-resolution strategy essentially utilizes multiple structured data containers at different resolution to disperse scene features at different scales. The higher-resolution structured data container brings a significant number of parameters, resulting in excessively large model sizes. Additionally, tri-plane, as an efficient structured data container, can reduce the storage pressure significantly by reducing three-dimensional spatial coordinate to three sets of two-dimensional data. However, the commonly employed axis-aligned mappings can lead to an imprecise representation of features. Our novel Gaussian derivation strategy addresses the limitations of axis-aligned mapping by dispersing sub-features of each primitive, achieving excellent rendering performance even at fine scales without relying on a multi-resolution mechanism.

\section{Preliminary of 3D Gaussian Splatting}
3D Gaussian Splatting \cite{3d_gaussian} utilizes anisotropic 3D Gaussian primitives to explicitly represent the underlying structure of the scene. The structure of each Gaussian is determined by two parameters defined in world coordinates: position (mean) $\bm{x}$ and 3D covariance matrix $\bm{\Sigma}$, 
\begin{equation}
    G(\bm{x}, \bm{\Sigma})=e^{-\frac{1}{2}\bm{x}^{T}\bm{\Sigma}^{-1}\bm{x}}.
\end{equation}
The intractable covariance matrix can be further decomposed into a scaling matrix $\bm{S}$ and a rotation matrix $\bm{R}$, where the correlation between them is $\bm{\Sigma}=\bm{R}\bm{S}\bm{S}^T\bm{R}^T$. For ease of optimization, we optimize a scaling vector $\bm{s}$ for $\bm{S}$ and a unit quaternion $\bm{q}$ for $\bm{R}$ in practice. Accordingly, the Gaussian is re-written as $G(\bm{x}, \bm{q}, \bm{s})$. In order to realize rendering, 3D Gaussians need to be projected onto the 2D image plane. Accordingly, the covariance matrix in camera coordinates can be obtained by $\bm{\Sigma}^{\prime}=\bm{J}\bm{W}\bm{\Sigma} \bm{W}^{T}\bm{J}^{T}$, where $\bm{W}$ denotes the view transformation matrix and $\bm{J}$ represents the Jacobian matrix approximating the projective transformation \cite{ewa_splatting,surface-splatting}. On the other hand, the appearance of every Gaussian is affected by the other two parameters, namely the opacity $\alpha$ and spherical harmonic coefficients $Y_{lm}$, which are combined with the spherical harmonic basis to represent view-dependent color. For each pixel on the image plane, its color is calculated by blending $N$ ordered 3D Gaussians above it:
% \begin{equation}
%     C=\sum_{i\in N}\bm{c}_{i}\alpha_i^{\prime}\prod_{j=1}^{i-1}(1-\alpha_j^{\prime}),
%     \label{equ3}
% \end{equation}
\begin{equation}
    C=\sum_{i\in N}\bm{c}_{i}\alpha_i\prod_{j=1}^{i-1}(1-\alpha_j),
    \label{equ3}
\end{equation}
where $\bm{c}_{i}$ represents the color of each Gaussian relying on the view direction, and $\alpha_i$ is computed by multiplying the learned opacity of the 3D Gaussian with the probability value of the corresponding 2D Gaussian projected onto the image plane at the current pixel. During the training of those Gassuians, they undergo alternant densification according to the conditions of under- or over-reconstruction and pruning primitives that do not contribute to the rendering results or have overlarge sizes. 

% $\alpha_i^{\prime}$ is obtained by multiplying the opacity $\alpha$ of the 3D Gaussian with its projected covariances $\Sigma^\prime$.

% \textcolor{blue}{where $\bm{c}_{i}$ and $\alpha_{i}$ are the color and opacity at the center of the 3D Gaussian multiplied by the probability value of its projection onto the two-dimensional plane as a 2D Gaussian distribution. Specifically, their relationship follows the formula below:}
% \textcolor{blue}{here, $c_{i}$ is the color of the i-th Gaussian individual overlapping the current pixel, $\alpha_{i}$ is derived from the 2D covariance and is influenced by the opacity in logit form $o_{i}$. The standard sigmoid function $\mathrm{sigma}(\cdot)$ is used to compute the opacity. Specifically, $\alpha_{i}(P)$ is given by:}

% \begin{equation}
%     \alpha_{i}(P)= \mathrm{sigm}(o_{i})\exp(-\frac{1}{2}(P-\mu_{i})^{T}(\Sigma_{i})^{-1}(P-\mu_{i})),
% \end{equation}
% \textcolor{blue}{in which, $\mu_{i}$ represents the coordinates of the center of the 3D Gaussian} During the training of those Gassuians, they undergo alternant densification according to the conditions of under- or over-reconstruction and pruning primitives that do not contribute to the rendering results or have overlarge sizes. 

% pipeline
\begin{figure*}[t] \centering
    \includegraphics[width=1\textwidth]{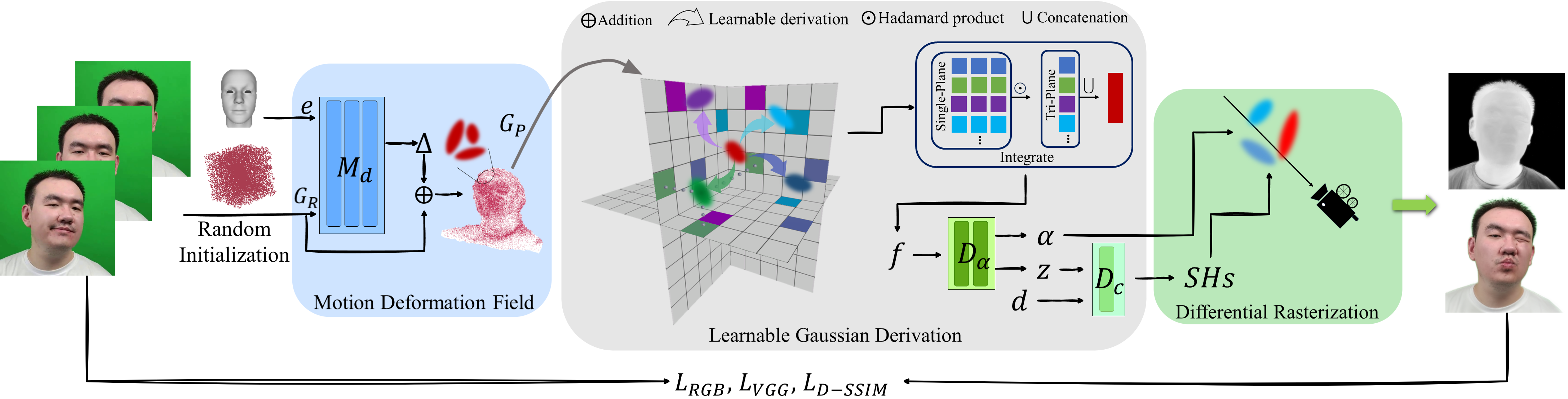}
    \caption{\textit{Method overview.} GaussianHead uses a set of 3D Gaussians with learnable attributes controlling their shape and appearance to model the subject's head. A motion deformation field is first set up to represent the dynamic head geometry, which converts structureless Gaussians $G_R$ to structured core ones $G_P$ in a posed space via conditioning on pre-acquired expression parameters $\bm{e}$. Next, a single-resolution tri-plane structure of the feature container is leveraged to store appearance-related attributes. Notably, derivation mechanisms through learnable rotations are applied to each core Gaussian, yielding several doppelgangers of it. The integration of sub-features obtained through projection onto the planes from those doppelgangers is taken as the final feature $\bm{f}$ of the core Gaussian. Two separate tiny MLPs are then employed to decode opacity $\alpha$ and spherical harmonic coefficients (SHs), based on which we generate the final rendering via differential rasterization.} 
    \label{fig:pipeline}
\end{figure*}
\section{Approach}
% \subsection{Overview}
Given a monocular video of the target person performing free head and facial movements, the corresponding expression parameters and masks of each frame, and known camera parameters, the proposed GaussianHead is trained on these four kinds of input. Our method models the complex head geometry based on anisotropic 3D Gaussians. A single-resolution tri-plane with a learnable derivation strategy is leveraged to encode texture information for those Gaussian primitives. Their integration ensures photorealistic rendering results. In inference, the approach allows high-fidelity reconstruction of the target subject and controllable animation by adjusting expression coefficients and view directions. Fig. \ref{fig:pipeline} illustrates the overall workflow.

\subsection{Motion Deformation Field}
The original 3D Gaussian splatting technique was devised to reconstruct static scenes. However, human heads in our dynamic setting exhibit complex motions. Thus, we construct a motion deformation field to obtain variable Gaussians connected with the subject's head motions, which are changeable within a period. Specifically, it first randomly initializes $M$ Gaussians $G_R = \{G(\bm{x}_i, \bm{q}_i, \bm{s}_i)\}_{i=1}^{M}$ in the initialized canonical space, each of which is connected with three geometric attributes: position $\bm{x}$, rotation represented by unit quaternion $\bm{q}$, and scaling vector $\bm{s}$. Subsequently, the pre-acquired facial expression parameters $\bm{e}$ and the positional encoding results of each Gaussian's spatial coordinate $\bm{x}$ are concatenated and then fed into a deformation network $M_d$ with a multi-layer perceptron (MLP) as its implementation. For a certain Gaussian, the deformation network predicts ${\Delta}_{\bm{x}}$, ${\Delta}_{\bm{q}}$, and ${\Delta}_{\bm{s}}$, which are offsets in terms of position, rotation, and scale:
\begin{equation}
    ({\Delta}_{\bm{x}},{\Delta}_{\bm{q}},{\Delta}_{\bm{s}})=M_{d}(\gamma(\bm{x}),\bm{e}),
\end{equation}
where $\gamma$ represents conducting position encoding on the mean position of the current 3D Gaussian, generating a high-dimensional sine-cosine sequence \cite{nerf}. Next, we add the obtained offsets to their initial states, yielding $\acute{\bm{x}}_i$, $\acute{\bm{q}}_i$, and $\acute{\bm{s}}_i$. Relevant updates are called core Gaussians $G_P = \{G(\acute{\bm{x}}_i, \acute{\bm{q}}_i, \acute{\bm{s}}_i)\}_{i=1}^{M}$ in the posed space.

% % comparision
% \begin{figure*}[htbp] 
% \centering
%     \includegraphics[width=1\textwidth]{image/sota_compar_v2.pdf}
%     \\
%     \caption{Qualitative comparisons of the reconstruction task. All competitors were run under the configurations specified by their respective works. Our method achieves superior visual results, particularly in aspects such as wrinkles, teeth, eyebrows, and even reflections on glasses.} 
%     \label{fig:sota_compar}
% \end{figure*}

\subsection{Learnable Gaussian Derivation} 
\label{sec:gaussian derivation}
\subsubsection{How to realize it?}
A single motion deformation MLP is sufficient to express complex head movements, but it struggles to simultaneously represent all fine and intricate facial textures (see Sec. \ref{sec:ablation}). To enhance the representation capability while using minor consumption, a \textit{single-resolution tri-plane} is employed to encode texture information around the subject's head. These factorized feature planes $\mathbf{P}_{xy}$, $\mathbf{P}_{xz}$, $\mathbf{P}_{yz}$ have an identical size of $(H, W, L)$, where $H$ and $W$ represent the height and width and $L$ denotes the channel length. Then, we derive $K$ doppelgangers from each core Gaussian by imposing an equivalent number of learnable derivations parameterized by unit quaternion $\bm{r}$ on it. In practice, we first generate the scalar and vector parts of the quaternion randomly and then do normalization to ensure its magnitude is 1. Since both the rotation axis and rotation angle are unfixed, the rotated Gaussian (i.e., doppelganger) is also variable in both the orientation and position. Note that $K$ is required to evenly divide $L$. For every individual doppelganger, it is vertically projected onto three factor planes, generating three individual plane feature vectors by sequentially interpolating $L/K$ channels of factor planes. Three shortened feature vectors from the tri-plane are fused by the Hadamard product, forming the sub-feature of the target doppelganger. Eventually, features from $K$ doppelgangers are concatenated, yielding the final feature representation $\bm{f}$ for the core Gaussian. Mathematically, that can be expressed as 
\begin{equation}
    \bm{f} = \bigcup\prod_{k=1}^K\prod_{j=1}^3\varphi(\mathbf{P}_j,\bm{r}_k(G_P)),
\end{equation}
where $\mathbf{P}$ represents the parameterized plane with $j$ indexing the $xy$, $xz$, and $yz$ factorized planes. The symbol $\bm{r}$ denotes the derivation transformation, which is optimized also via a unit quaternion in our practice. In addition, $\varphi$ indicates the projection of a doppelganger onto a factor plane and subsequent bilinear interpolation, and $\bigcup$ means concatenating short features from multiple doppelgangers.
\par

The unit quaternions controlling the derivation do not have an explicit first-order gradient. Due to manifold constraints, they cannot be straightforwardly optimized in Euclidean space using first-order optimizers. Therefore, we employ the Riemannian ADAM optimizer \cite{riemannian_adam}, where updates to unit quaternions at the training step $t$ are induced by an exponential term based on the learning rate $\alpha_t$ and gradient $\bigtriangledown_t$ in their tangent space:
\begin{equation}
    \bm{r}_{k,t+1} = \bm{r}_{k,t} \mathrm{Exp}(\alpha_t \bigtriangledown_t).
\end{equation}

\subsubsection{Why does it work?}
An often overlooked fact is that when the tri-plane with axis-aligned mapping is used for feature representation, the three-dimensional space defined by these orthogonal planes simultaneously couples geometric with appearance attributes. Specifically, any point within this space serves a dual purpose: it not only represents the local shape structure of the object to be modeled but also determines locations on three feature planes for storing the corresponding local appearance information. Regarding the former, in order to achieve local watertightness and completeness of the object's structure, the point primitives used to represent the target object are typically very dense and occupy only a small portion of the three-dimensional space constructed by the tri-plane. For the latter, when these densely distributed points are projected onto the three feature planes in an axis-aligned manner, the projection regions only cover a small fraction of the entire feature plane, leading to low storage space utilization. Additionally, this results in the first type of feature dilution problem described in the Introduction section (i.e., multiple points being too close together, causing their projections to coincide in the same cells across the three feature grids). Although the multi-resolution mechanism employed by the tri-plane can somewhat alleviate this first type of feature dilution problem, it introduces a large number of parameters. Furthermore, the storage space at each resolution level still remains underutilized, as the projected points only occupy a small subset of the cells on the feature planes. As for the second type of feature dilution problem (i.e., multiple points being located along the same line perpendicular to a feature plane), it cannot be alleviated by this mechanism. Fig.~\ref{fig:explain-dilution}(a) visually illustrates the feature dilution issue, aiding comprehension of the preceding discussion.
\par

Our method selects 3D Gaussians as the representation primitives, which, to some extent, can also be viewed as points. The core reason why our proposed “derivation strategy” is effective lies in its decoupling of the two functions of the points in the three-dimensional space corresponding to the tri-plane: one function for representing the object’s geometry and the other for determining the storage location of the appearance features. In practice, we create multiple doppelgangers for each core Gaussian and then use them to determine the storage location for the feature information associated with the current Gaussian point. In other words, we assign the core Gaussian to represent the geometric attributes of the modeled object, while its corresponding doppelgangers are responsible for acquiring the corresponding appearance attributes. This approach effectively avoids the representation ambiguity caused by storing the features of multiple Gaussians in the same location. An additional benefit of this method is that, theoretically, we can create doppelgangers at any location in the entire three-dimensional space corresponding to the tri-plane, thereby achieving full utilization of the storage region. To gain a deeper understanding of how our design mitigates the feature dilution problem, please refer to Fig.~\ref{fig:explain-dilution}(b).

\subsection{Hierarchical Radiance Decoding and Inherited Derivation Initialization}
In contrast to recent dynamic scene approaches based on 3D Gaussians \cite{rencent_gaussian_work1,rencent_gaussian_work2} that directly set opacity and color as optimizable parameters, we employ two small MLPs to decode the final features into opacity $\alpha$ and spherical harmonic coefficients $Y_{lm}$: 
\begin{equation}
    (\alpha, \bm{z}) = D_{\alpha}(\bm{f}),
\end{equation}
\begin{equation}
    Y_{lm} = D_{c}(\bm{d},\bm{z}),
\end{equation}
where $\bm{z}$ denotes the intermediate latent code and $\bm{d}$ is the view direction. We use 4th-order spherical harmonic coefficients to synthesize view-dependent colors. This design ensures a more precise inference of texture (validated in Sec. \ref{sec:ablation}). Finally, differentiable rasterization in Eq. \ref{equ3} is used for rendering.
\par

Recall that the number of Gaussian primitives dynamically increases or decreases during their training process. Given a new Gaussian copied by or split from the original entity, it proceeds to derive multiple doppelgangers via optimizable rotations in our method. Regarding the optimization of rotation parameters, we employ an inheritance initialization strategy, which means initializing them with the same values the parent Gaussian uses. Experiments show that, compared to random or zero initialization, the inheritance strategy increases the training speed by approximately $25\%$. We describe this setting in more detail and conduct experimental comparisons in Sec. \ref{sec:init method}.

\subsection{Training Objectives}
We utilize L1 loss $L_{\mathrm{RGB}}$ to measure the pixel-wise difference between the ground truth image and its corresponding rendering. The perceptual loss $L_{\mathrm{VGG}}$ \cite{lpips} and D-SSIM loss $L_{\mathrm{D-SSIM}}$ are leveraged to measure the quality loss of the rendered images. Specifically, for the perceptual loss, we use the first four layers of the VGG model \cite{vgg} to extract feature maps from compared image pairs for subsequent similarity computation. The full objective function as well as its components are given as follows:
\begin{equation}
    L = \lambda_1 L_{\mathrm{RGB}} + \lambda_2 L_{\mathrm{VGG}} + \lambda_3 L_{\mathrm{D-SSIM}},
\end{equation}
\begin{equation}
    L_{\mathrm{RGB}}=||\mathbf{I}-\mathbf{I}^{\mathrm{GT}}||_1,
\end{equation}
\begin{equation}
    L_{\mathrm{VGG}}=||\mathrm{VGG}(\mathbf{I})-\mathrm{VGG}(\mathbf{I}^{\mathrm{GT}})||_1,
\end{equation}
\begin{equation}
    L_{\mathrm{D-SSIM}}=1-\mathrm{SSIM}(\mathbf{I}, \mathbf{I}^{\mathrm{GT}}),
\end{equation}
where we empirically set $\lambda_1 = 0.8$, $\lambda_2 = 0.01$, and $\lambda_3 = 0.2$. 

\newcommand{\tabincell}[2]{\begin{tabular}{@{}#1@{}}#2\end{tabular}}

\section{Experiments}

\subsection{Datasets and Baselines}
All monocular videos for experiments are sourced from public subjects, including real-life and internet recordings. We take the first part of every video clip for models' training, varying between 2000 and 2500 frames. The remaining ones constitute the testing samples. For each subject, we pre-process corresponding videos to get four modality signals: RGB head images with a unified resolution of 512$\times$512, expression parameters tracked by a 3DMM model \cite{bfm}, camera parameters, and binary masks predicted by MODNet \cite{modnet}. Regarding the head movement, we follow the practice in \cite{nerface,nerfblendshape}, which anchors the head in the coordinate system and simulates its pose changes with camera poses.

We compare our approach with six state-of-the-art methods for evaluation purposes. SplattingAvatar \cite{SplattingAvatar} binds 3D Gaussian to 3DMM, thus capable of utilizing the excellent rendering quality of 3D Gaussian and the effective motion control manner of 3DMM to construct a head avatar. INSTA \cite{INSTA} is based on volumetric NeRF and reconstructs avatars by building a neural surface on the foundation of 3DMM. PointAvatar \cite{pointavatar} is based on explicit points and constructs detailed avatars through iterative refining and upsampling point clouds. NeRFBlendShape \cite{nerfblendshape} constructs a hybrid NeRF and utilizes a multi-resolution hash grid to store facial expression bases. MonoGaussianAvatar\cite{monogaussianavatar} models dynamic head avatars by leveraging 3D Gaussian points in combination with a Gaussian deformation field, and introduces a novel insertion-deletion strategy to facilitate improved convergence. GaussianBlendShape\cite{gaussianblendshape} represents the head avatar as a linear combination of a base model of neutral expression and a set of expression blendshapes, all of which are represented by learned 3D Gaussians. It also imposes a consistency constraint between the Gaussian blendshapes and the mesh blendshapes, significantly reducing artifacts in the boundary regions of the subject under novel expressions.
% represents head avatars using a linear combination of a neutral expression-based Gaussian model and a set of expression and their blendshapes represented by 3D Gaussians.

\subsection{Implementation Details}
We implement GaussianHead with PyTorch, where differentiable 3D Gaussian rasterization is implemented based on CUDA kernels \cite{3d_gaussian}. The initialization included 10$K$ 3D Gaussians. Apart from optimizing the derivative quaternion using Riemannian ADAM \cite{riemannian_adam}, all other optimizers are first-order ADAM \cite{adam}. Specifically, for the unit quaternion $\bm{q}$ representing 3D Gaussian rotation, we follow the explicitly first-order gradients as derived in \cite{3d_gaussian} and also optimize it using ADAM. In our experiments, we set the batch size to 1 and train the model on a single RTX 3090, taking approximately 2 hours. After 3000 iterations, we perform densification and pruning on 3D Gaussians every 500 iterations, pruning Gaussians larger than a certain proportion of the scene and removing Gaussians with opacity below the threshold. We set the size threshold for large Gaussians to be greater than 1\% of the scene and the opacity threshold to 0.0002. Other details follow the settings in \cite{3d_gaussian}.

For the tri-plane structural feature container, we set the size of each plane to $64\times 64 \times 32$ ($H\times W \times L$). Regarding the derivative unit quaternions $\bm{r}$, they are treated as trainable parameters, initialized with random numbers in the range [0, $2^{32}$], and uniformly distributed on a 3-dimensional hypersphere to obtain unit quaternions with better distribution properties. Their learning rate decays from $1\times10^{-3}$ to 0 at the last iteration.

% comparision
\begin{figure*}[htbp] 
\centering
    \includegraphics[width=1\textwidth]{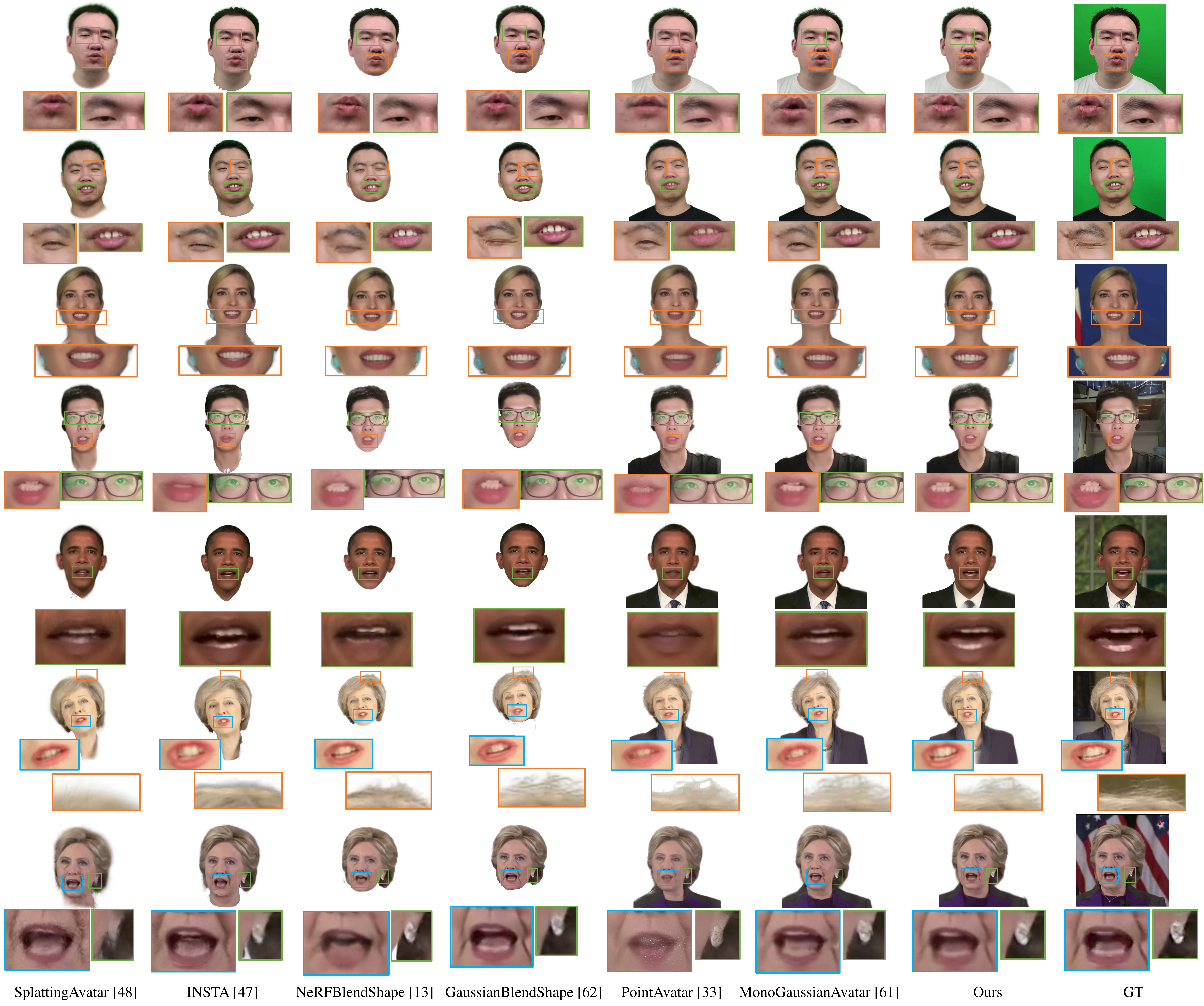}
    \\
    \caption{Qualitative comparisons of the reconstruction task. All competitors are run under the configurations specified by their respective works. Our method achieves superior visual results, particularly in aspects such as wrinkles, teeth, eyebrows, and even reflections on glasses.} 
    \label{fig:sota_compar}
\end{figure*}
\subsection{Comparison with State-of-the-art Approaches}
\textit{Reconstruction.} This task is to construct a lifelike avatar of the target person in the reference video and make it perform the same actions. We showcase visual comparison results rendered by our method and baselines in Fig. \ref{fig:sota_compar}. PointAvatar uses the built-in linear blend skinning (LBS) algorithm of 3DMM to control facial movements. By linearly combining expression coefficients and expression bases, this limited representation method often fails to accommodate extreme expressions. As evidence, please see the snapshots of avatars generated by this approach in Fig. \ref{fig:sota_compar}, which fail to accurately depict mouth movements. Dynamic results in the supplementary video present a more intuitive demonstration of this particular deficiency. In addition, since the primitives adopted are points with fixed shapes, this approach may yield holes during reconstruction ($7$th row). 
SplattingAvatar employs 3DMM mesh as the geometry prior of the human head and makes all Gaussians attached to triangle meshes. With this design, in practice, it realizes animation akin to INSTA, which is based on direct mesh movement rather than LBS techniques. However, a side effect is that it needs to calculate the variation of 3D Gaussian properties using the motion of coarse mesh triangles, which occasionally results in inaccurate modeling of facial expressions. NeRFBlendShape exhibits rendering blur when reconstructing complex surfaces such as lips and eyes, which may be due to the inaccurate density estimation in the modeling of geometric surfaces. INSTA models the head based on a predefined 3DMM geometry, making it struggle to represent the fine structures (e.g., the jewelry in the $3$rd and the suspended hair in the $6$th row). MonoGaussianAvatar relies on the rough FLAME template and skinning weights and uses the LBS algorithm for motion control, and thus performs relatively poorly in capturing subtle expressions (such as the eye-closing action in the 2nd row). 
% GaussianBlendShape struggles to accurately reproduce subtle movements in certain frames (e.g., the slight opening of the mouth in the fifth and sixth rows), possibly due to its reliance on precise learning of expression blendshapes. 

% Additionally, the linear combination of multiple blendshapes may amplify errors in certain micro-expressions.

In contrast, our GaussianHead can accurately render subtle details such as slightly closed eyes and wrinkles. Even if only a few frames show the oral cavity in the training data, we can achieve precise reconstruction (refer to Fig. \ref{fig:ablate_dif}). According to the best of our knowledge, this achievement is hard to obtain for previous head avatar approaches that also train their algorithms on monocular videos. 
%error map
\begin{figure}[t] 
\centering
    \includegraphics[width=0.5\textwidth]{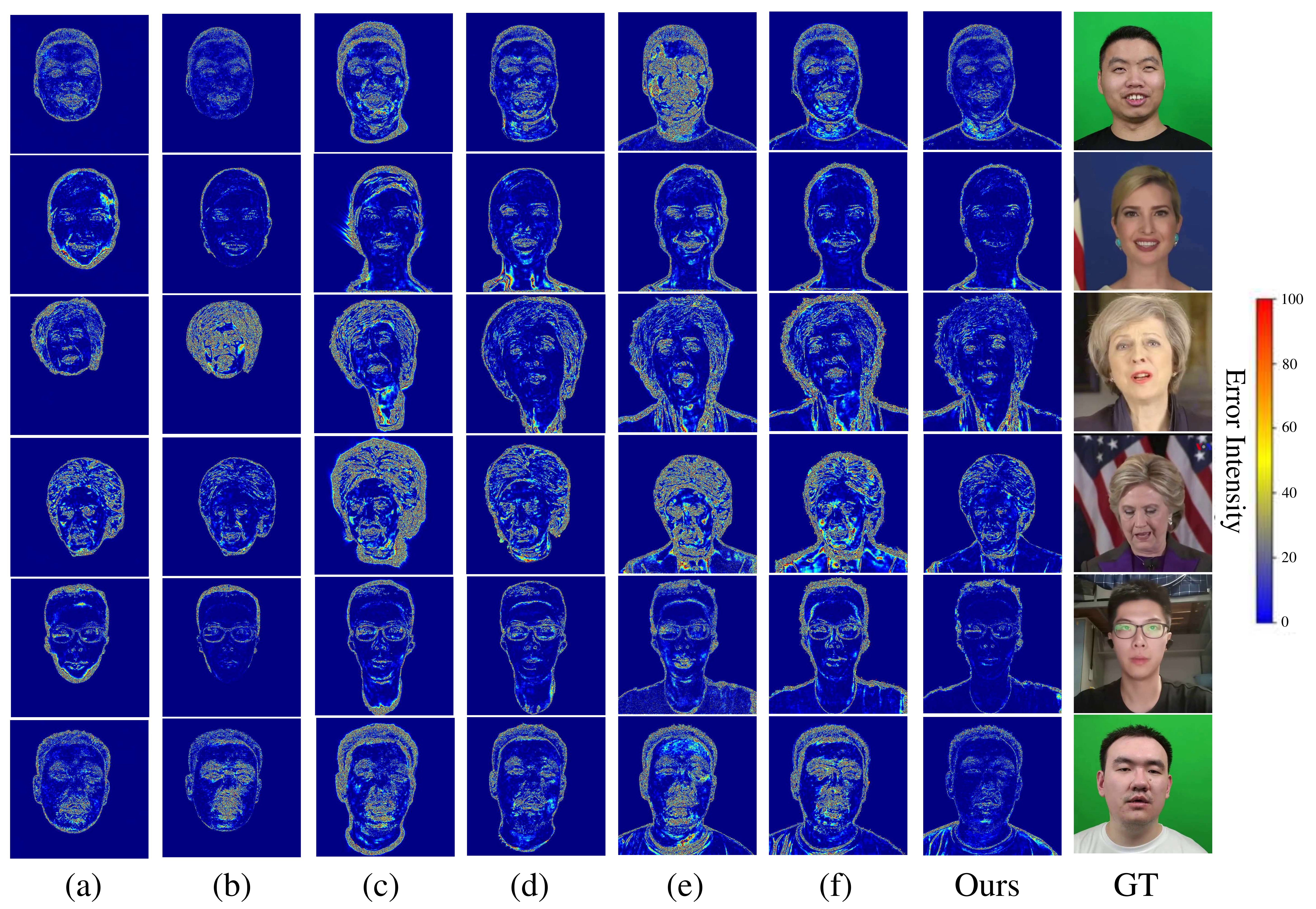}\\
    \caption{Error maps for the reconstruction task. We compare with (a) NeRFBlendShape\cite{nerfblendshape}, (b) GaussianBlendShape \cite{gaussianblendshape}, (c) SplattingAvatar \cite{SplattingAvatar}, (d) INSTA \cite{INSTA}, (e) PointAvatar \cite{pointavatar} and (f) MonoGaussianAvatar \cite{monogaussianavatar}. Note that methods (a-d) only model the head, and (e-f), as well as ours, further include the torso. In each map, brighter areas indicate larger errors.} 
    \label{fig:error-map}
\end{figure}

\begin{table*}[ht]
\caption{Quantitative comparisons of reconstruction quality. The six subjects in Fig. \ref{fig:sota_compar} are indexed as ID1-ID6 from top to bottom.}
\label{tab:metric}
\centering
\begin{threeparttable}[b]
\begin{tabular}{ll|lll|lll|lll}
\hline
 &  &  & \multicolumn{1}{c}{ID1} &  &  & \multicolumn{1}{c}{ID2} & &  & \multicolumn{1}{c}{ID3} & \\
\multicolumn{1}{c}{}   & \multicolumn{1}{c|}{Method} & \multicolumn{1}{c}{PSNR$\uparrow$} & \multicolumn{1}{c}{SSIM$\uparrow$} & \multicolumn{1}{c|}{LPIPS$\downarrow$} & \multicolumn{1}{c}{PSNR$\uparrow$} & \multicolumn{1}{c}{SSIM$\uparrow$} & \multicolumn{1}{c|}{LPIPS$\downarrow$} & \multicolumn{1}{c}{PSNR$\uparrow$} & \multicolumn{1}{c}{SSIM$\uparrow$} & \multicolumn{1}{c}{LPIPS$\downarrow$} \\ \hline
\multirow{5}{*}{\rotatebox[origin=c]{90}{w/o torso}} 
& INSTA\cite{INSTA}  & 27.41  &0.892  & 0.141  &28.43  &0.906  &0.172 &25.26  & 0.903 &0.189 \\
& SplattingAvatar\cite{SplattingAvatar}  & 28.42  &0.901   & 0.119   &27.24   & 0.873  & 0.181 &26.14  & 0.915  & 0.186 \\
& NeRFBlendShape\cite{nerfblendshape}  &30.93  &0.913 & 0.117   & 30.12 & 0.917 & 0.124  & 28.92  &\underline{0.961}  &  \underline{0.085}  \\
& GaussianBlendShape\cite{gaussianblendshape}  & \textbf{32.42}  & \underline{0.922}   & \underline{0.111}  & \textbf{31.26} & \underline{0.938} & \textbf{0.105} & \underline{29.96} & 0.936  & 0.115  \\
& GaussianHead (Ours\textsuperscript{*}) & \underline{31.85} & \textbf{0.927} & \textbf{0.098}   & \underline{30.75} & \textbf{0.941}  & \underline{0.108}  &\textbf{30.05}   & \textbf{0.972}  & \textbf{0.061}  \\ \hline
\multirow{3}{*}{\rotatebox[origin=c]{90}{w/ torso}}   
& PointAvatar\cite{pointavatar}  & 26.76  & 0.861 & 0.162 &  23.78 & 0.803 & 0.191 & 24.61 & 0.794 & 0.182  \\
& MonoGaussianAvatar\cite{monogaussianavatar}  & \underline{29.13}  & \underline{0.916}   & \underline{0.106}  & \underline{29.07}  & \underline{0.909}  & \underline{0.127}   & 27.45  & \underline{0.947}  & \underline{0.093}  \\
& GaussianHead (Ours)  &\textbf{31.70}   &\textbf{0.940}  & \textbf{0.091}   & \textbf{29.21}  & \textbf{0.911}  & \textbf{0.119}  & \textbf{28.70} &  \textbf{0.963}  & \textbf{0.066}  \\ \hline
 & & & \multicolumn{1}{c}{ID4} & & \multicolumn{1}{c}{ID5} &  &  & \multicolumn{1}{c}{ID6}  & \\
\multicolumn{1}{c}{} & \multicolumn{1}{c|}{Method} & \multicolumn{1}{c}{PSNR$\uparrow$} & \multicolumn{1}{c}{SSIM$\uparrow$} & \multicolumn{1}{c|}{LPIPS$\downarrow$} & \multicolumn{1}{c}{PSNR$\uparrow$} & \multicolumn{1}{c}{SSIM$\uparrow$} & \multicolumn{1}{c|}{LPIPS$\downarrow$} & \multicolumn{1}{c}{PSNR$\uparrow$} & \multicolumn{1}{c}{SSIM$\uparrow$} & \multicolumn{1}{c}{LPIPS$\downarrow$} \\ \hline
\multirow{5}{*}{\rotatebox[origin=c]{90}{w/o torso}} 
& INSTA\cite{INSTA}   &29.79  & 0.931  &0.095  & 26.77   &0.791  & 0.131   & 27.19   & 0.876  & 0.095   \\
& SplattingAvatar\cite{SplattingAvatar}  &30.28   &0.928  &0.099  & 26.78  & 0.760    & 0.113  & 26.25  &  0.812   & 0.113   \\
& NeRFBlendShape\cite{nerfblendshape}   &30.26  & 0.952  &  0.094   & 27.61   & 0.837   & \underline{0.081}  & 27.09   & \underline{0.916}  & \underline{0.075}  \\
& GaussianBlendShape\cite{gaussianblendshape}  &\underline{30.73} &\underline{0.956}   & \underline{0.082}   & \underline{28.26}  & \underline{0.929}   & 0.159   & \underline{27.86}  & 0.909  &0.124    \\
& GaussianHead (Ours\textsuperscript{*})   &\textbf{30.97}   & \textbf{0.969}  & \textbf{0.076}   & \textbf{28.73}   & \textbf{0.962}  & \textbf{0.032}  & \textbf{28.52}  & \textbf{0.973}   & \textbf{0.030}   \\ 
\hline
\multirow{3}{*}{\rotatebox[origin=c]{90}{w/ torso}}   
& PointAvatar\cite{pointavatar} &27.11   & 0.898   &  0.142   &  23.36  & 0.642   & 0.154  & 24.71   & 0.717  & 0.185  \\
& MonoGaussianAvatar\cite{monogaussianavatar}  &\underline{30.02}  & \underline{0.921}  &\underline{0.097}  & \underline{27.01}  & \underline{0.841}  & \underline{0.102}  & \underline{27.10}  &  \underline{0.899}   & \underline{0.098}   \\
& GaussianHead (Ours)   &\textbf{30.06}  & \textbf{0.947}   &\textbf{0.089}  &\textbf{27.65}  & \textbf{0.893}  & \textbf{0.095}   & \textbf{28.31}   & \textbf{0.904}  & \textbf{0.084}   \\ 
\hline
\end{tabular}
\begin{tablenotes}
    \item[1] The number with bold typeface means the best result and the underline is the second best.
    \item[2] For a fair comparison, all image metrics were tested under the condition of no background rendering. To compare with those methods without considering the torso part, we also report our corresponding version by removing the torso, denoted by a superscript symbol ``*".
\end{tablenotes}
\end{threeparttable}
\end{table*}

On the other hand, the error map, as opposed to the RGB image, can reflect the reconstruction accuracy more clearly. In practice, we calculate the L1 distance between the reconstructed images and the corresponding ground truth at the pixel level, the results of which are further mapped into the RGB space to produce the final error maps. Fig. \ref{fig:error-map} clearly shows that our method achieves better reconstruction accuracy than others. Evaluation results tabulated in Table \ref{tab:metric} quantitatively demonstrate our superiority under three image quality related metrics.

% % reenact
%  \begin{figure*}[htbp] \centering
%     \includegraphics[width=1\textwidth]{image/reenact_v2.pdf}
%     \caption{Qualitative comparisons of the cross-identity reenactment task. Here, (a), (b), (c), and (d) respectively denote NeRFBlendShape \cite{nerfblendshape}, PointAvatar \cite{pointavatar}, INSTA \cite{INSTA}, and SplattingAvatar \cite{SplattingAvatar}. Our GaussianHead achieves the best reenactment results, even in conveying extreme expressions. For more intuitive comparisons of motion sequences reenacted by these methods, please refer to the supplementary video.} 
%     \label{fig:reenact}
% \end{figure*}
\textit{Reenactment.} This setting is more challenging than the reconstruction task because it aims to transfer the motion patterns of one real subject to the constructed avatar with a different identity. We showcase the reenactment results in Fig. \ref{fig:reenact}. MonoGaussianAvatar, PointAvatar and SplattingAvatar struggle to reproduce even non-subtle and non-extreme expressions. In some frames, significant distortions in poses or expressions are observed. For illustration, in the 1st row of Fig. \ref{fig:reenact}, MonoGaussianAvatar and PointAvatar fails to reproduce the neck's state accurately. In the second row, SplattingAvatar cannot model the entire face successfully. The biggest problem with INSTA and {GaussianBlendShape} is the apparent identity discrepancy with the animated avatar where the source and target human objects have significantly different external shapes (e.g., one person has a long face, and the other has a round face). In other words, the avatar and its corresponding real human object look like belonging to two different persons. For evidence, please see the snapshots by them two in the 2nd and 5th rows of Fig. \ref{fig:reenact}. Note that this issue does not show up in the reconstruction results. NeRFBlendShape performs less satisfactorily in depicting facial details, as all the rendering frames appear over-smoothed. By comparison, the head avatars animated by GaussianHead excel at precisely reproducing the actions and poses of the source subject while depicting facial details and preserving identity consistency simultaneously. Our method still performs well even in imitating extreme facial expressions—for example, the first and third rows of Fig. \ref{fig:reenact}.
% reenact
 \begin{figure*}[htbp] \centering
    \includegraphics[width=1\textwidth]{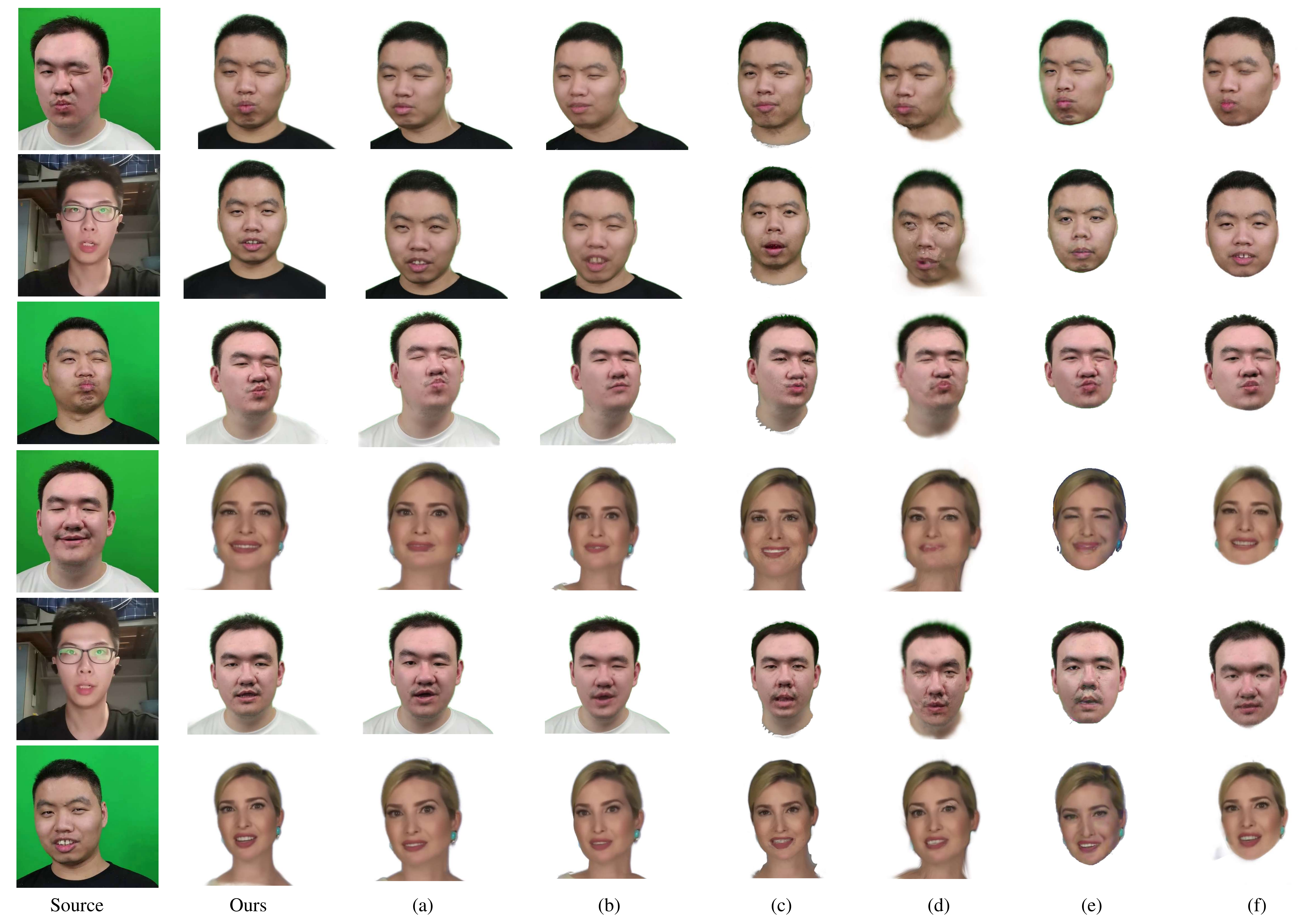}
    \caption{Qualitative comparisons of the cross-identity reenactment task. Comparison methods include (a) MonoGaussianAvatar\cite{monogaussianavatar}, (b) PointAvatar\cite{pointavatar}, (c) INSTA\cite{INSTA}, (d) SplattingAvatar \cite{SplattingAvatar}, (e) GaussianBlendShape \cite{gaussianblendshape} and (f) NeRFBlendShape \cite{nerfblendshape}. Our GaussianHead achieves the best reenactment results, even in conveying extreme expressions. For more intuitive comparisons of motion sequences reenacted by these methods, please refer to the supplementary video.} 
    \label{fig:reenact}
\end{figure*}

\begin{figure}[t] 
\centering
    \includegraphics[width=0.5\textwidth]{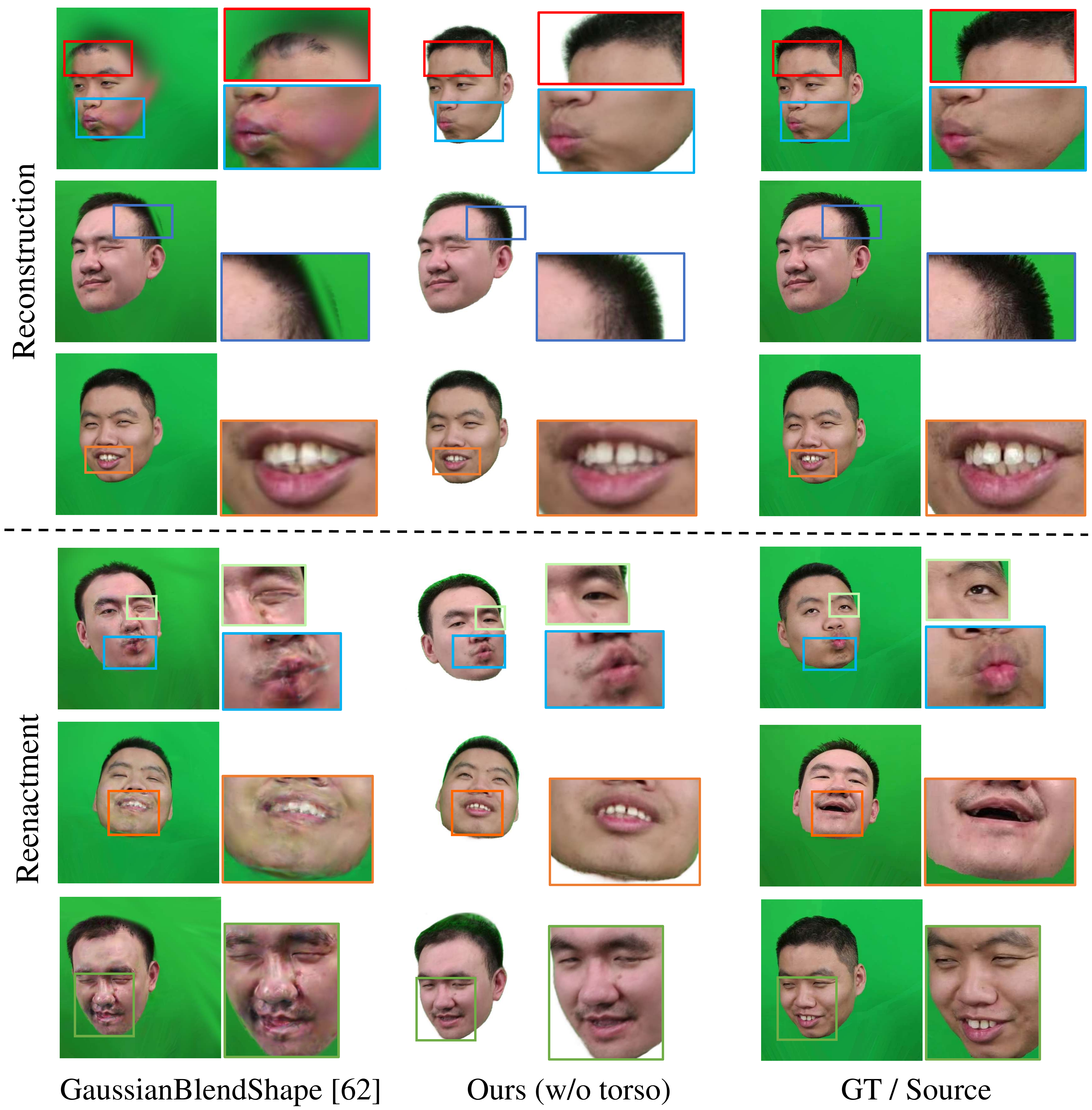}\\
    \caption{Comparative results with GaussianBlendShape on challenging cases.} 
    \label{fig:challenge_case}
\end{figure}

\emph{Fine comparison with GaussianBlendShape.} While GaussianBlendShape demonstrates comparable performance to our method in less challenging scenarios, particularly when subjects are in near-frontal views as presented in Figs. \ref{fig:sota_compar} and \ref{fig:reenact}, we conduct further evaluations on them, focusing on more extreme conditions to assess their performance ceiling rigorously. Representative examples are presented in Fig. \ref{fig:challenge_case} and the supplementary video. These results lead to the following observations: 1) GaussianBlendShape exhibits limitations in accurately reconstructing poses where the subject is in profile or near-profile views; 2) In cross-identity reenactment, GaussianBlendShape occasionally produces inaccuracies, including incorrect motion replication, blurred rendering of fine-grained facial details, and color artifacts in the lower face. This is expected, as the reenactment task poses greater challenges than reconstruction. It requires the constructed avatar to replicate the motions and poses of a driving actor with a different identity while maintaining high image quality in the rendered video frames. Insufficient modeling and disentanglement of the involved facial and head attributes will lead to degraded results; 3) Our GaussianHead demonstrates robust performance in these challenging scenarios, with significant advantages over GaussianBlendShape, especially in the reenactment setup. Furthermore, as highlighted in Table \uppercase\expandafter{\romannumeral 2}, our method offers a substantial advantage in model size, being approximately $66\times$ smaller (12MB vs. 800MB) than GaussianBlendShape. This significant reduction in model size makes our approach particularly well-suited for deployment on resource-constrained devices.

\textit{More evaluations.} Apart from rendering quality, we also conduct evaluations in multiple aspects and record the results in Table \ref{tab:model_size}. In brief, our method consumes the minimum GPU memory at inference time, smaller training resource and medium rendering speed with respect to other methods. It is especially noteworthy that we have a much smaller model size (\textit{$1/2$ of the second smallest}) than competitors. It should be attributed to the derivation mechanism adopted in our framework, which allows the precise encoding of appearance information of 3D Gaussians with a single-resolution tri-plane-formed feature container. More detailed investigation of that will be given in Section \ref{sec:ablation}.

%indices model size
\begin{table}[tbp]
\caption{Quantitative comparisons of model size, training cost, and rendering time.}
\label{tab:model_size}
\centering
\begin{threeparttable}[b]
    \label{tab:table1}
    \resizebox{0.48\textwidth}{!}{
    % \large
    \begin{tabular}{*{2}{l|cccc}}
        \toprule
       Method   & \makecell{Model \\ Size (MB)$\downarrow$} & \makecell{Rendering Time\\ (second)$\downarrow$} & \makecell{Train. \\ Memory (GB)$\downarrow$} & \makecell{Infer. \\Memory (GB)$\downarrow$}\\
        \midrule
        SplattingAvatar \cite{SplattingAvatar}  & 84 & \underline{0.04} & 11 & 5\\
        INSTA \cite{INSTA}  &\underline{60} & 0.06 & 16 & 3.5\\
        NeRFBlendShape \cite{nerfblendshape}  & 564 & 0.28 &\textbf{7.5} & \underline{3}\\
        PointAvatar \cite{pointavatar}  & 212 & 1.5 & 80  & 23\\
        % \midrule
        GaussianBlendShape\cite{gaussianblendshape} & 800 &\underline{0.04} & 12 & 4.5 \\
        MonoGaussianAvatar\cite{monogaussianavatar}  &25.7  &\textbf{0.03}  &10.5  &7.5 \\
        GaussianHead (\textbf{ours})  & \textbf{12} & 0.09 &\underline{8.5} & \textbf{2.5}\\
        \bottomrule
    \end{tabular}
    }
    \begin{tablenotes}
        \item [1] The number with bold typeface means the best and the underline is \\
        the second best.
        \item [2] The average rendering time of a single frame is tested on an RTX3090.
    \end{tablenotes}
    \end{threeparttable}
\end{table}

% novel view
\begin{figure*}[thb] \centering
   \includegraphics[width=1\textwidth]{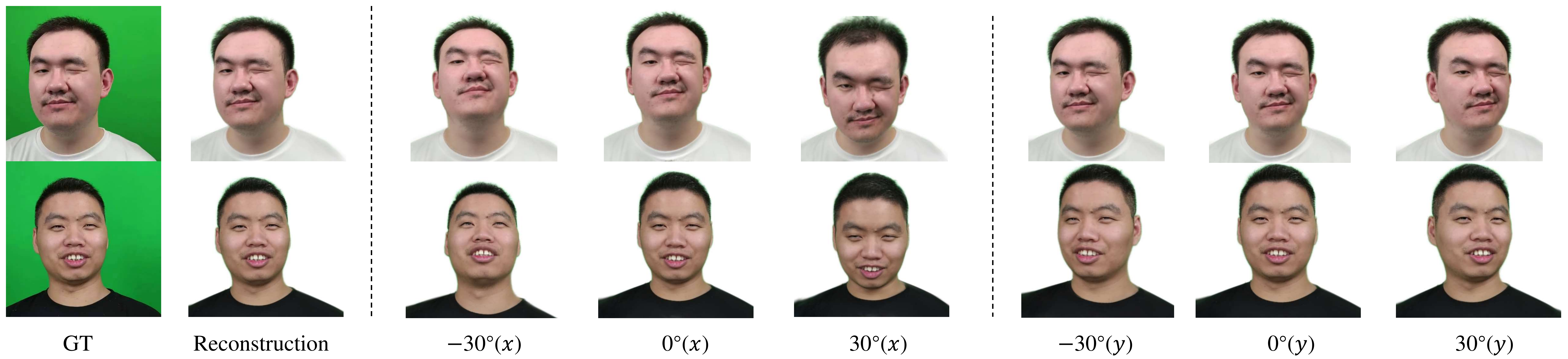}
    \caption{Novel views rendered by our GaussianHead via rotating the camera viewpoint around the $x$ or $y$ axes. There exists a pronounced consistency in rendering details across these newly synthesized perspectives.} 
    \label{fig:novel_view}
\end{figure*}
\subsection{Novel View Synthesis and Depth Estimation}
Avatars generated by the proposed GaussianHead exhibit remarkable multi-view consistency, as illustrated in Fig. \ref{fig:novel_view}. The results show no artifacts or unrealistic facial expressions. At the same time, our method can also accurately represent details, such as teeth, from novel perspectives. Based on the settings of previous work \cite{rencent_gaussian_work2}, we additionally present the reconstructed precise depth maps in Fig. \ref{fig:depth}, from which it can be clearly seen whether there is the presence of artifacts. Our method demonstrates excellent depth estimation performance.
%depth
\begin{figure}[tbp] 
\centering
    \includegraphics[width=0.48\textwidth]{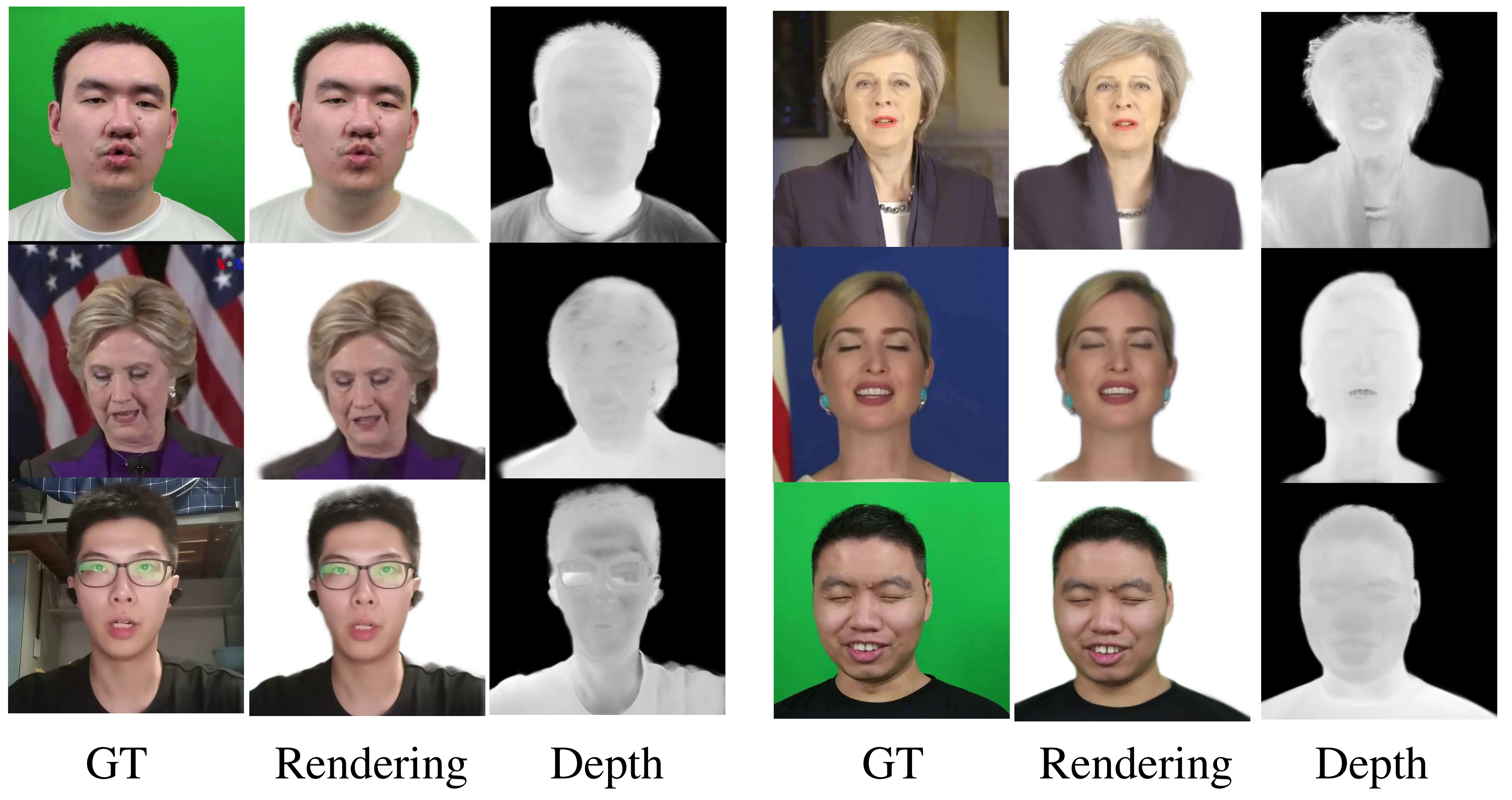}
    \caption{
    We reconstruct the ground truth image (left) into a head avatar (center) and simultaneously visualized the depth estimated by GaussianHead (right). Our method obtains accurate depth information.} 
    \label{fig:depth}
\end{figure}

%T-experiment
\begin{figure}[tbp]
  \centering
   \includegraphics[scale=0.2]{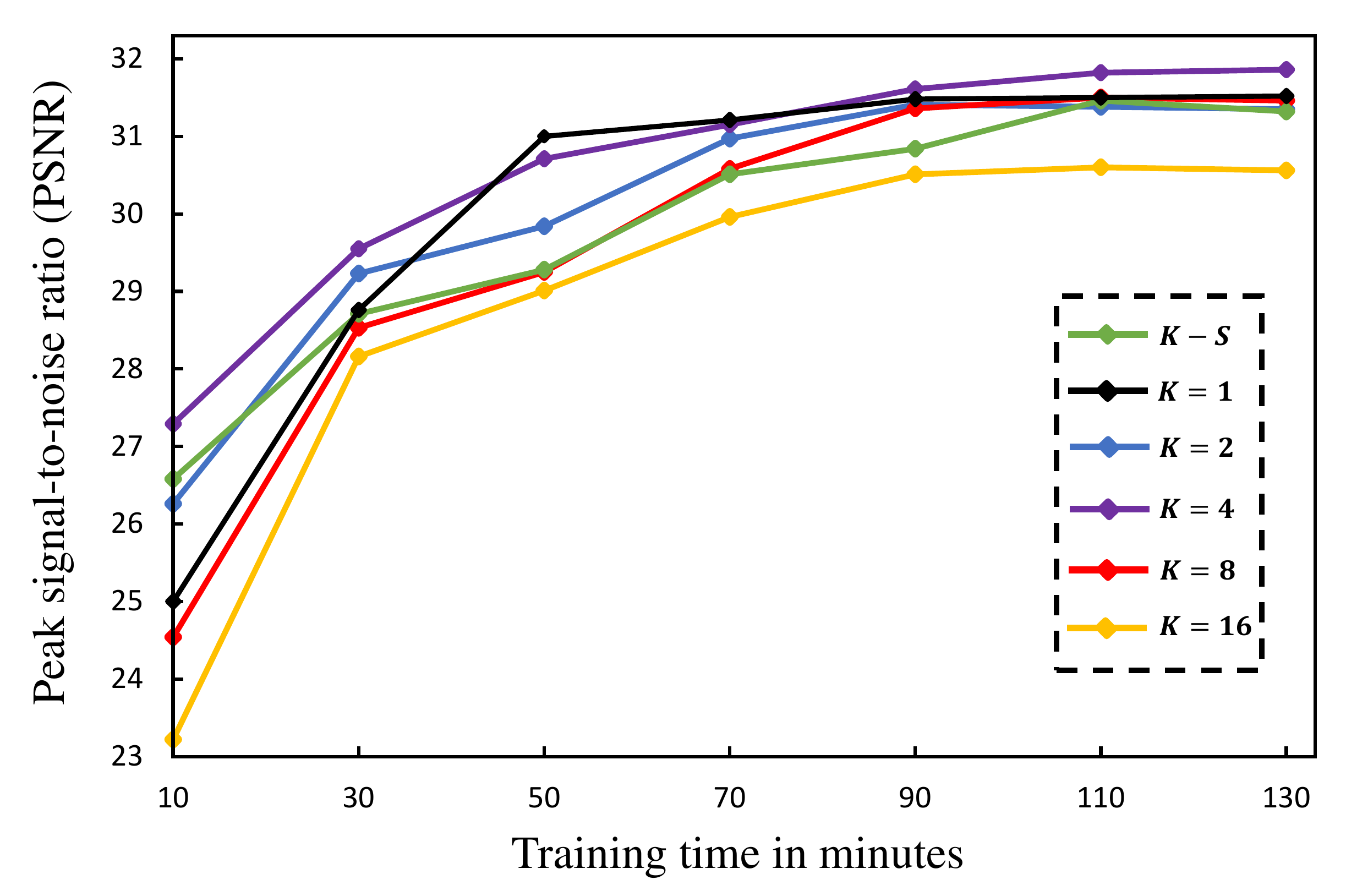}
   \caption{The performance of taking different numbers of derived doppelgangers $K$ during the training process (``$K-S$" represents the $K$-scheduler). Optimal performance is achieved when setting $K$ to 4.}
    \label{fig:T_experiment}
\end{figure}

% %ablation multi-resolution
% \begin{table}[tbp]
% \caption{Performance of using different resolution mechanisms for tri-plane and different initialization strategies for Gaussian derivation.}
% \centering
%     \resizebox{0.48\textwidth}{!}{
%     \large
%     \begin{tabular}{*{10}{l|ccccc}}
%         \toprule
%          Method  & PSNR$\uparrow$ &  SSIM$\uparrow$ & LPIPS$\downarrow$  & model size & training time\\
%         \midrule
%         M-res   & 29.89 & 0.931 & 0.090 & 136 MB & $\sim$ 3 hours \\
%         \midrule
%         R-init &28.85 & 0.912 & 0.103 &  7.5 MB & $\sim$ 2.5 hours\\
%         Z-init &28.42 & 0.897 & 0.105 &  7.5 MB & $\sim$ 2.5 hours\\
%         \midrule
%         Ours (S-res, I-init) & 29.87 & 0.938 & 0.092 &  7.5 MB & $\sim$ 2 hours\\
%         \bottomrule
%     \end{tabular}
%     }
%     \label{tab:multi-res}
% \end{table}

\subsection{Number of Doppelgangers}
The number of derived doppelgangers ($K$) significantly influences training time and the final rendering results. We conduct experiments to determine the optimal value by setting it to five different values: 1, 2, 4, 8, and 16 (under the requirement that the count of channels $L$ in each factor plane, set to 32 in our implementation, is divisible by $K$). Additionally, we also investigate an adaptive $K-$scheduler for a more comprehensive evaluation: increasing $K$ with the growing number of Gaussians:
\begin{equation}
    K=\frac L{2^{t-u}},  u=\left\lceil\frac{n_i}{n_o}\right\rceil,
\end{equation}
where $n_i$ and $n_o$ are the current and initialized number of Gaussians, $\left\lceil\cdot\right\rceil$ denotes the ceiling function, and $t$ is a constant set to 6. Relevant experimental results are shown in Fig. \ref{fig:T_experiment}. In the context of a fixed value, an excessively large $K$ results in optimization challenges, notably evidenced by a pronounced deceleration in the optimization process at $K=8$ and $16$. Conversely, fewer doppelgangers ($K=1, 2$) are proven to be inadequate in addressing the issue of feature dilution, leading to a discernible deterioration in PSNR. On the other hand, the incorporation of an adaptive changing scheme (i.e., $K-$scheduler) does not yield superior outcomes, which should be partially attributed to the predominant occurrence of newly introduced Gaussians in vacant regions. These regions, in contrast to their densely optimized counterparts, have a lower likelihood of encountering feature dilution problems. According to the experimental results, we ultimately choose $K=4$. 

\subsection{Initialization of Derivative Quaternion }
\label{sec:init method}
In neural radiance fields \cite{nerf,tilted}, the number of sampled points in the scene is mostly fixed. However, our GaussianHead undergoes densification or pruning of 3D Gaussians as the iterations progress. For the Gaussians that are clipped out, their quaternions used for controlling derivation are directly deleted. For the newly added ones that split from the parent Gaussian, three initialization schemes for derivative quaternions are studied: randomly initializing from the range [0, $2^{32}$] as done at the beginning of training (R-init), zero initialization (Z-init), and inheriting derivative parameters from the parent Gaussian (I-init). The time required for these three schemes to achieve their respective optimal rendering quality is reported in Table \ref{tab:multi-res}. We observe that the inheriting initialization leads to better rendering results and quicker convergence than its two counterparts. We think these advantages stem from the relatively good starting values provided by the parent Gaussians, reducing the difficulty of parameters optimization.
%ablation multi-resolution
\begin{table}[tbp]
\caption{Performance of using different resolution mechanisms for tri-plane and different initialization strategies for Gaussian derivation.}
\label{tab:multi-res}
\centering
    \resizebox{0.48\textwidth}{!}{
    % \large
    \begin{tabular}{*{10}
    {l|cccccc}}
    % \begin{tabular}{*{10}{l|ccccc}}
        \toprule
         Method  & PSNR$\uparrow$ &  SSIM$\uparrow$ & LPIPS$\downarrow$  & \makecell{Model Size \\ (MB)$\downarrow$} & \makecell{Training \\ Time (hours)$\downarrow$} & \makecell{Rendering \\ Time (second)$\downarrow$}\\
        \midrule
        M-res   & 29.89 & 0.931 & 0.090 & 136 & $\sim$ 3 & 0.15\\
        \midrule
        R-init &28.85 & 0.912 & 0.103 &  12 & $\sim$ 2.5 & 0.09\\
        Z-init &28.42 & 0.897 & 0.105 &  12 & $\sim$ 2.5 & 0.09\\
        \midrule
        {Ours (S-res, I-init)} & 29.87 & 0.938 & 0.092 &  12 & $\sim$ 2 & 0.09\\
        \bottomrule
    \end{tabular}
    }
\end{table}

% %ablation study
% \begin{table}[tp]
% \caption{Ablation studies on several key components of our GaussianHead.}
% \centering
%     \begin{threeparttable}[b]
%     \resizebox{0.48\textwidth}{!}{
%     \large
%     \begin{tabular}{*{10}{l|cccc}}
%         \toprule
%          Method&  L1$\downarrow$ & PSNR$\uparrow$ &  SSIM$\uparrow$ & LPIPS$\downarrow$  \\
%         \midrule
%         w/o tri-plane  & 0.0014 & 28.53 & 0.891 & 0.144  \\
%         w/o derivation  & 0.0013 & 28.89 & 0.912 & 0.115  \\
%         w/o ${\Delta}_{\bm{s}}$ &- &- &- &- \\
%         w/o ${\Delta}_{\bm{q}}$ &0.0015 &28.25 &0.901 &0.099 \\
%         w/o ${\Delta}_{\bm{s}}$ and ${\Delta}_{\bm{q}}$ &- &- &- &- \\
%         w/o perceptual loss &0.0017 &27.82 &0.897 &0.151 \\
%         Direct deform &0.0035 &24.55 &0.792 &0.319 \\
%         Global rotation & 0.0015 & 28.31 & 0.901 & 0.110 \\
%         Decoded by MLP &0.0023 &26.42 &0.871 &0.195 \\
%         \midrule
%         Ours full  &\textbf{0.0010}&\textbf{29.87} &\textbf{0.938} &\textbf{0.092} \\
%         \bottomrule
%     \end{tabular}
%     }
%    \begin{tablenotes}
%      \item[1] The leading dash (-) means the model dose not converge.
%    \end{tablenotes}
%   \end{threeparttable}
%     \label{tab:ablation-indices}
% \end{table}

%ablate triplane
\begin{figure}[t] 
\centering
    \includegraphics[width=0.48\textwidth]{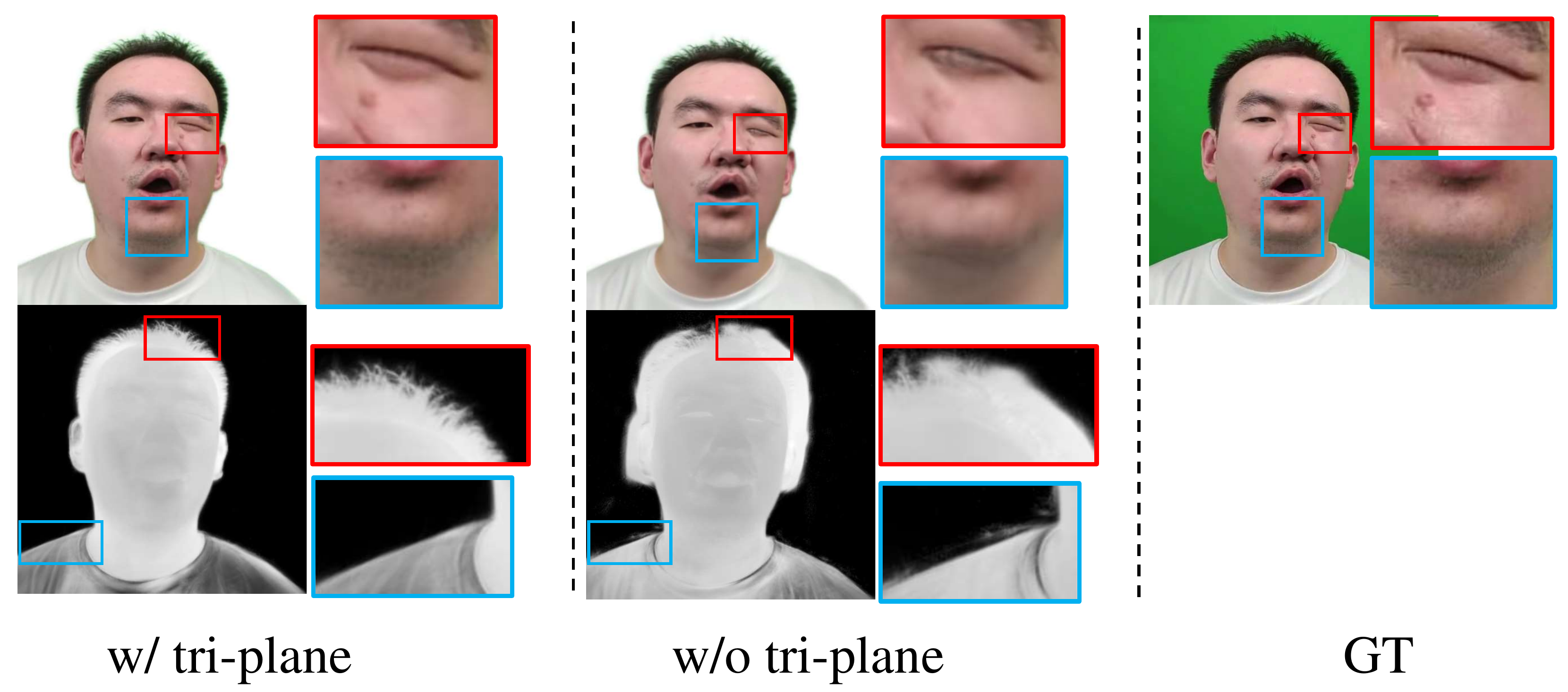}
    \\
    \caption{Visual comparisons of with (w/) or without (w/o) employing the parametric tri-plane structure to store Gaussians appearance information.}
    \label{fig:ablate_plane}
\end{figure}

% %main ablate
% \begin{figure*}[tbp] 
% \centering
%     \includegraphics[width=1\textwidth]{image/main_ablate.pdf}
%     \\
%     \caption{Qualitative demonstration of partial ablation experiments. The novel Gaussian derivation strategy accurately restores features representing various complex structure regions, enabling the high-fidelity rendering of subtle details even for oral cavity rarely seen in the training data.}
%     \label{fig:ablate_dif}
% \end{figure*}

% %ablate vgg
% \begin{figure}[t] 
% \centering
%     \includegraphics[width=0.45\textwidth]{image/direct_deform.pdf}
%     \\
%     \caption{Visual results without employing perceptual loss and using direct deformation field.}
%     \label{fig:ablate_vgg}
% \end{figure}

\subsection{Ablation Study}
\label{sec:ablation}
Unless otherwise stated, we conduct ablation experiments on the first three subjects and present the average results.
%ablation study
\begin{table}[tp]
\caption{Ablation studies on several key components of our GaussianHead.}
\label{tab:ablation-indices}
\centering
    \begin{threeparttable}[b]
    \resizebox{0.48\textwidth}{!}{
    % \large
    % \fontsize
    \footnotesize
    \begin{tabular}{*{10}{l|cccc}}
        \toprule
         Method&  L1$\downarrow$ & PSNR$\uparrow$ &  SSIM$\uparrow$ & LPIPS$\downarrow$  \\
        \midrule
        w/o tri-plane  & 0.0014 & 28.53 & 0.891 & 0.144  \\
        w/o derivation  & 0.0013 & 28.89 & 0.912 & 0.115  \\
        w/o ${\Delta}_{\bm{s}}$ &- &- &- &- \\
        w/o ${\Delta}_{\bm{q}}$ &0.0015 &28.25 &0.901 &0.099 \\
        w/o ${\Delta}_{\bm{s}}$ and ${\Delta}_{\bm{q}}$ &- &- &- &- \\
        w/o perceptual loss &0.0017 &27.82 &0.897 &0.151 \\
        Direct deform &0.0035 &24.55 &0.792 &0.319 \\
        Global rotation & 0.0015 & 28.31 & 0.901 & 0.110 \\
        Decoded by MLP &0.0023 &26.42 &0.871 &0.195 \\
        \midrule
        Ours full  &\textbf{0.0010}&\textbf{29.87} &\textbf{0.938} &\textbf{0.092} \\
        \bottomrule
    \end{tabular}
    }
   \begin{tablenotes}
     \item[1] The leading dash (-) means the model dose not converge.
    \item [2] The number with bold typeface means the best and the underline is the second best.
   \end{tablenotes}
  \end{threeparttable}
\end{table}

\noindent\textit{Parametric tri-plane.}
Our method employs a parametric tri-plane as the container to store Gaussian appearance information. For comparison, we ablate the parametric tri-plane and directly set the opacity and spherical harmonic coefficients connected with 3D Gaussians as optimizable parameters (``w/o tri-plane"). The ablation results given in Fig. \ref{fig:ablate_plane} validate the effectiveness of our design. The depth maps yielded by directly optimizing these two radiance parameters lead to numerous redundant Gaussians around the head that will not be removed because their opacity value is above the pre-defined threshold. This inevitably increases training overhead and causes negative perceptual impacts (pay attention to the enlarged local area on the face). 

Furthermore, to demonstrate the advantages of parameterized tri-plane in compactly and accurately encoding texture-related knowledge, we replace it with a multi-layer perceptron (``Decoded by MLP") having a comparable number of parameters. Regarding this variant, our pipeline consists of two MLPs, one for the motion deformation field and the other for decoding texture semantic information from posed Gaussians. When checking its results in Fig. \ref{fig:ablate_dif}, we find that in region not commonly seen in the training videos, such as the oral cavity, it usually renders facial components in colors different from the ground truth. In contrast, our GaussianHead with tri-plane gives precise texture reproduction, even like the lip color and the highlight above earrings. Quantitative results summarized in Table \ref{tab:ablation-indices} align with this observation.
%main ablate
\begin{figure*}[tbp] 
\centering
    \includegraphics[width=1\textwidth]{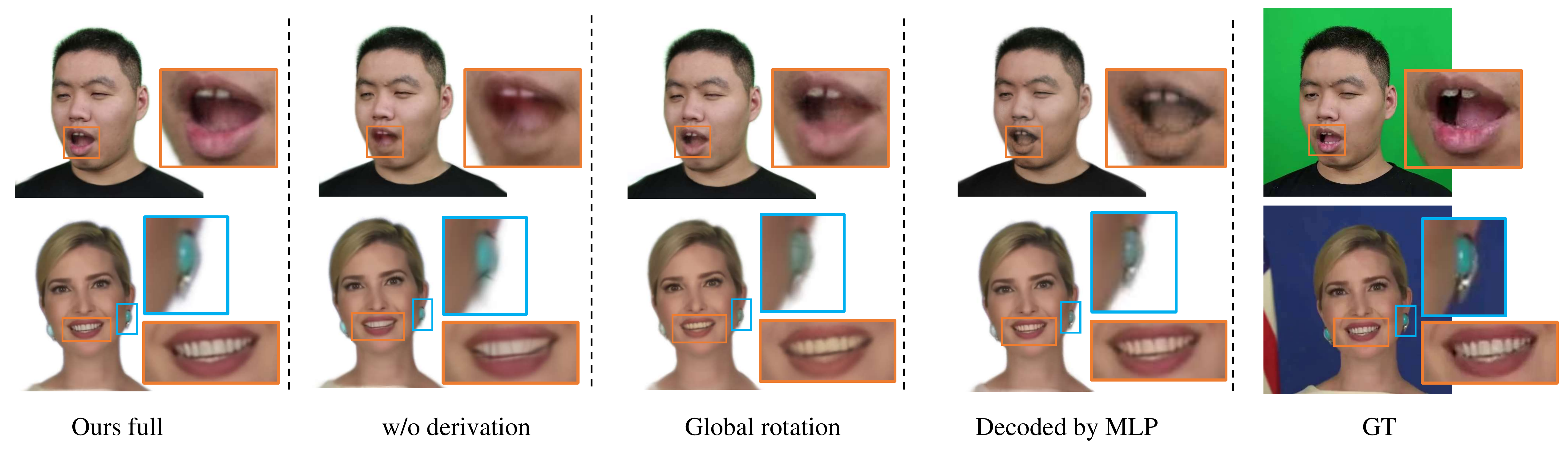}
    \\
    \caption{Qualitative demonstration of partial ablation experiments. The novel Gaussian derivation strategy accurately restores features representing various complex structure regions, enabling the high-fidelity rendering of subtle details even for oral cavity rarely seen in the training data.}
    \label{fig:ablate_dif}
\end{figure*}

\noindent\textit{Learnable Gaussian derivation.} We make ablation experiments using axis-aligned mappings, following the practice in the traditional tri-plane \cite{eg3d,k-planes} (``w/o derivation"). Compared to our GaussianHead leveraging the derivation strategy, this variant performs worse in expressing high frequency signals. Additionally, we imitate TILTED \cite{tilted} by adopting a fixed set of rotations for the entire scene (``Global rotation") rather than individual primitives. The experimental results support our analysis in Sec. \ref{sec:introduction}, global rotation is unsuitable for complex dynamic scenarios.

\noindent\textit{Multi-resolution tri-plane.} In works like k-planes \cite{k-planes} and hexplane \cite{hexplane}, the multi-resolution mechanism of tri-plane is crucial for obtaining high-quality results because of the feature extraction at different scales. However, it also leads to an increase of model size. We investigate the impacts of using this design in our framework (following the settings of previous work \cite{k-planes}, we use multiple resolutions of $[64, 128, 256, 512]$ in this experiment). Relevant results in Table \ref{tab:multi-res} show that the induced improvement in rendering quality is almost negligible, whereas the model size grows dramatically. Meanwhile, utilizing a single-resolution tri-plane in our method already achieves comparable visual effects. This should be attributed to the proposed Gaussian derivation strategy, which allows the efficient and precise representation of facial appearance. 

% We devise an efficient Gaussian derivation mechanism to suppress the feature dilution problem among different Gaussian primitives, effectively preserving the unique feature attributes of each Gaussian. This allows us to use a tri-plane feature structure with single resolution to accurately encode texture-related properties. On the other hand, abandoning the multi-resolution mechanism significantly reduces the size of our model. We report the results in Tab. \ref{tab:multi-res}. It can be observed that using the multi-resolution mechanism brings little improvement in the rendering quality. Meanwhile, utilizing a single-resolution tri-plane and derivation mechanism achieves comparable results while significantly reducing the model size and training time cost. Considering both quality and performance, we opted to use a single-resolution tri-plane (Note that we consider the time it takes for the model to reach the highest quality metric as the training time).

\noindent\textit{Motion deformation field.} The motion deformation field is crucial to express dynamic-related expressions. It deserves further exploration on whether directly learning the deformed geometric attributes or predicting their offsets relative to current values. We start with our investigation by separately removing ${\Delta}_{\bm{q}}$ ("w/o ${\Delta}_{\bm{q}}$"), ${\Delta}_{\bm{s}}$ ("w/o ${\Delta}_{\bm{s}}$") and both of them ("w/o ${\Delta}_{\bm{q}}$ and ${\Delta}_{\bm{s}}$") to evaluate their influence. This means that during final rendering, we directly take the initialized rotation ${\bm{q}}$, scaling ${\bm{s}}$ or both of them as the geometric attributes of deformed 3D Gaussians in the posed space. Experiments show that after removing ${\Delta}_{\bm{s}}$, some Gaussians begin to fluctuate uncontrollably, either becoming too large or too small, leading to the failure of model convergence.

We push forward the ablation by using a MLP to directly predict the properties of 3D Gaussians in the posed space. In this setup, the deformation field predicts all geometric parameters of posed 3D Gaussians, rather than estimating relevant residual values and then adding them to the initials. We report relevant performance ("Direct deform") in Table \ref{tab:ablation-indices} and Figure \ref{fig:ablate_vgg}. 
The experimental results indicate that, both in terms of quantitative metrics and visual effects, predicting the geometric properties of posed Gaussians directly rather than their residual values yields significantly inferior results. We deduce that direct inference poses many difficulties to the network in specifically distinguishing head regions influenced by dynamic expression parameters, resulting in widespread blurring in areas outside the face.

% The former may be attributed to the network's difficulty in specifically distinguishing the head region influenced by dynamic expression parameters, resulting in widespread blurring in visual effects. When using residual connections, the residual values reflect which regions undergo larger dynamic changes and superimpose these changes onto the initial geometric properties. For regions with minimal changes, the residual values are optimized to be extremely small, and even if added, they would not affect the initial Gaussian's geometric properties. Therefore, for representing dynamic scenes, this separate handling of residual connections for static and dynamic regions is a preferable choice. 

\noindent\textit{Perceptual loss.} During the training process, our GaussianHead is supervised by minimizing a perceptual loss, actually striving to shorten the distance between the features extracted from two images using a pre-trained VGG \cite{vgg} model. We observe that employing this loss effectively preserves the consistency in details between the rendered image and its origin reference. By comparison, the removal of it causes a significant decrease in the rendering quality, as evidenced by the results in Fig. \ref{fig:ablate_vgg} and Table \ref{tab:ablation-indices}.
%ablate vgg
\begin{figure}[t] 
\centering
    \includegraphics[width=0.48\textwidth]{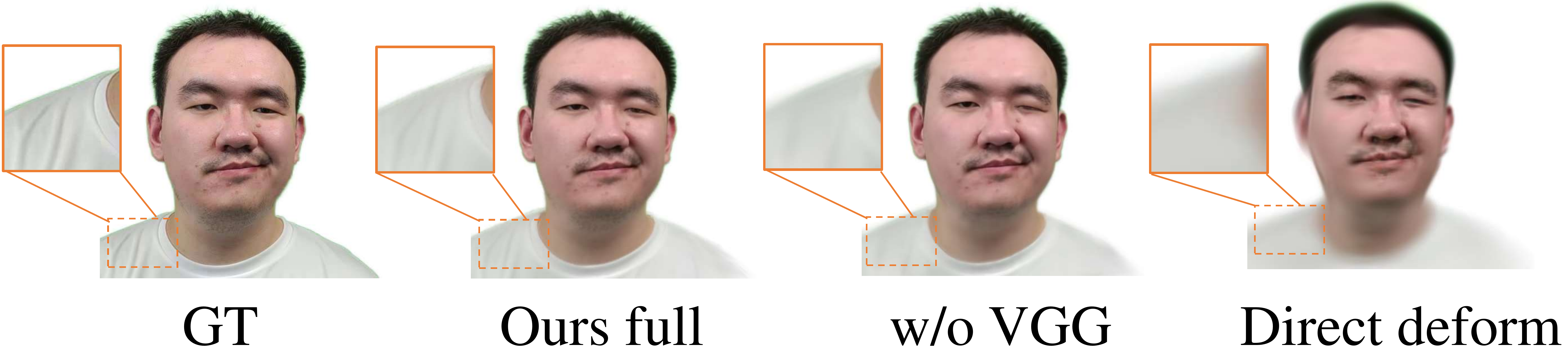}
    \\
    \caption{Visual results without employing perceptual loss and using direct deformation field.}
    \label{fig:ablate_vgg}
\end{figure}

% \section{More Details}
% \noindent\textbf{Dataset Acquisition} To facilitate a fair comparison, all our data is sourced from public subjects, including real-life and internet videos. The training data for each subject comprises approximately 2000 to 2500 frames, and the test dataset is conducted using the last 3\% to 5\% of frames. Specifically, the data for each subject we used includes only four parts: RGB head images with a resolution of 512$\times$512, expression parameters (derived from 3DMM model fitting \cite{bfm}), camera parameters, and binary masks created using MODNet \cite{modnet}. In terms of head movement, we anchored the head in the coordinate system and simulated head pose changes using camera poses \cite{nerface,nerfblendshape}.

%failure case
\begin{figure}[thp] 
\centering
    \includegraphics[width=0.48\textwidth]{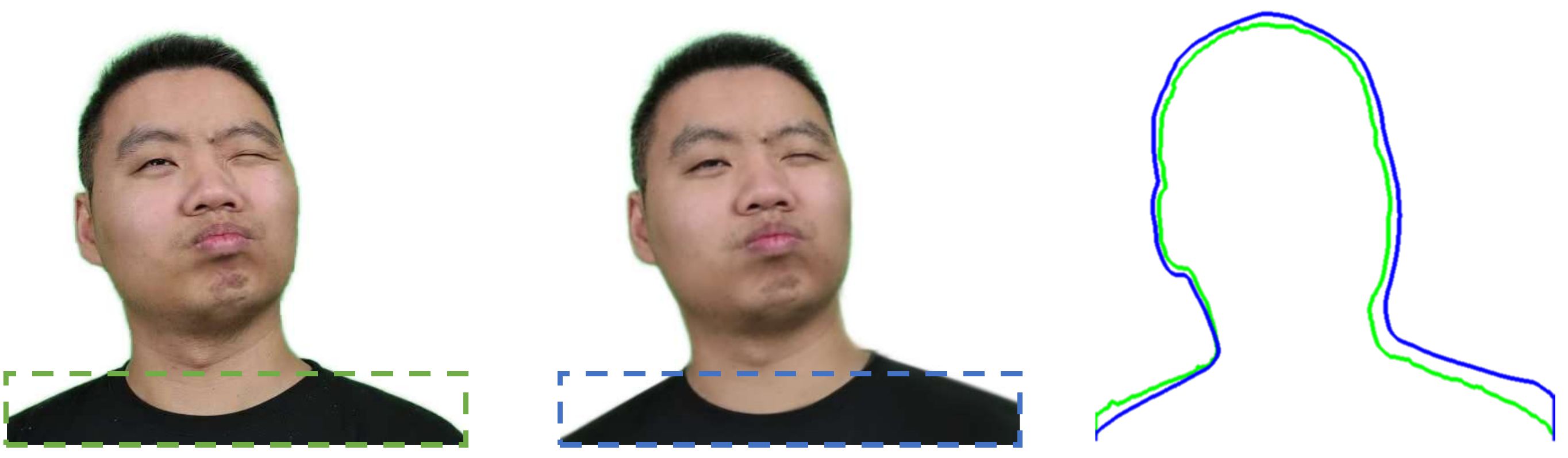}
    \\
    \caption{Failure case. When significant movement occurs in the ground truth (left), the rendered avatar's torso (middle) will exhibit visible corresponding movement. For clarity, we extract and superimpose their contours (right) for comparison. Compared to the ground truth contours (green), the contours of the rendered avatar (blue) show certain offsets in the torso. Please refer to the supplementary video for the animation.}
    \label{fig:limitation}
\end{figure}

\section{Discussion}
Although our method achieves excellent visual effects, it does not separate the motion of the head and torso. Instead, it still uses camera parameters as rough proxies for pose parameters, which is a common practice in some avatar works (e.g., NeRFBlendShape \cite{nerfblendshape} and INSTA \cite{INSTA}). However, this may result in unnatural shoulder shaking in scenarios with significant motion differences between the head and torso (see Fig. \ref{fig:limitation}). Existing separation methods either utilize semantic masks for independent training of these two parts \cite{adnerf,radnerf,er_nerf} or explicitly control their motions separately using the LBS algorithm of 3DMM \cite{imavatar,pointavatar}. The former complicates the entire training process, making it unable to achieve end-to-end training and incurring additional training overhead. The latter, while capable of separately controlling the broad-range pose movements of the head and torso, performs poorly in controlling subtle expressions (because it is difficult for 3DMM to represent most complex expressions), as demonstrated by the expression distortion issues in the supplementary video of PointAvatar. Therefore, exploring a practical end-to-end method that achieves separate control of partial structures on the head while effectively expressing subtle facial expressions should be a valuable direction for future work. Some promising solutions include:
\begin{itemize}
    \item Endowing each Gaussian with a new attribute to reveal which part it models—the head or the torso. Being aware of this, we can construct a deformation field purely responsible for facial dynamics and use the LBS to control the articulated changes in head and torso poses.
    % \item By assigning semantic attributes to each Gaussian and combining it with the advantage of LBS for accurately controlling poses and the ability of the motion deformation field to accurately represent expressions, to ensure that the expression changes controlled by the deformation field occur only on the face, while LBS is used to control the pose.
    \item  Incorporating a virtual joint approach similar to the one used in \cite{gart} to compensate for muscle movements not fully controlled by LBS, thereby addressing the limitations of 3DMM in expression control.
\end{itemize}

% %failure case
% \begin{figure}[thp] 
% \centering
%     \includegraphics[width=0.5\textwidth]{image/limitation.pdf}
%     \\
%     \caption{Failure case. When significant movement occurs in the ground truth (left), the rendered avatar's torso (middle) will exhibit visible corresponding movement. For clarity, we extracted and superimposed their contours (right) for comparison. Compared to the ground truth contours (green), the contours of the rendered avatar (blue) show certain offsets in the torso. Please refer to the supplementary video for the animation.}
%     \label{fig:limitation}
% \end{figure}

\section{Conclusion}
% This paper introduces the GaussianHead based on 3D anisotropic Gaussians, which can construct high-fidelity head avatars and achieve reliable animations. Technically, we present an effective idea of applying Gaussian primitives to model geometrically variable heads in continuous motions and represent complex textures. Due to the compact tri-plane feature structure and novel derivation strategy, our method demonstrates remarkable performance in handling challenging scenarios, including skin details, fluffy hair, oral cavities with scarce training data, and extreme expressions. At the same time, it maintains a tiny model size. We believe that GaussianHead will provide universal inspiration for followers in this field.
In conclusion, this paper introduces GaussianHead, a novel approach leveraging 3D anisotropic Gaussians to construct high-fidelity head avatars and produce reliable animations. Our method presents an effective framework for utilizing Gaussian primitives to model geometrically variable heads with continuous motions and represent complex textures. With its compact single-resolution tri-plane feature structure and innovative derivation strategy, GaussianHead demonstrates exceptional performance in handling challenging scenarios, such as intricate skin details, voluminous hair, limited training data for oral cavities, and extreme facial expressions, all while maintaining a considerably small model size. We anticipate that GaussianHead will serve as a valuable source of inspiration for the followers.

\section*{Acknowledgments}
This work was supported in part by the National Nature Science Foundation of China under Grants 61931012, 62301278, and 62371254; in part by the Science and Technology Development Fund, Macau SAR, under Grants 0141/2023/RIA2 and 0193/2023/RIA3; in part by the Nature Science Foundation of Jiangsu Province of China under Grants BK20230362 and BK20210594; and in part by the Natural Science Research Start-up Foundation of Recruiting Talents of Nanjing University of Posts and Telecommunications under Grants NY222019 and NY221019.


\bibliographystyle{IEEEtran}
\bibliography{uncolored_ref}

\begin{IEEEbiography}[{\includegraphics[width=1in,height=1.25in,clip,keepaspectratio]{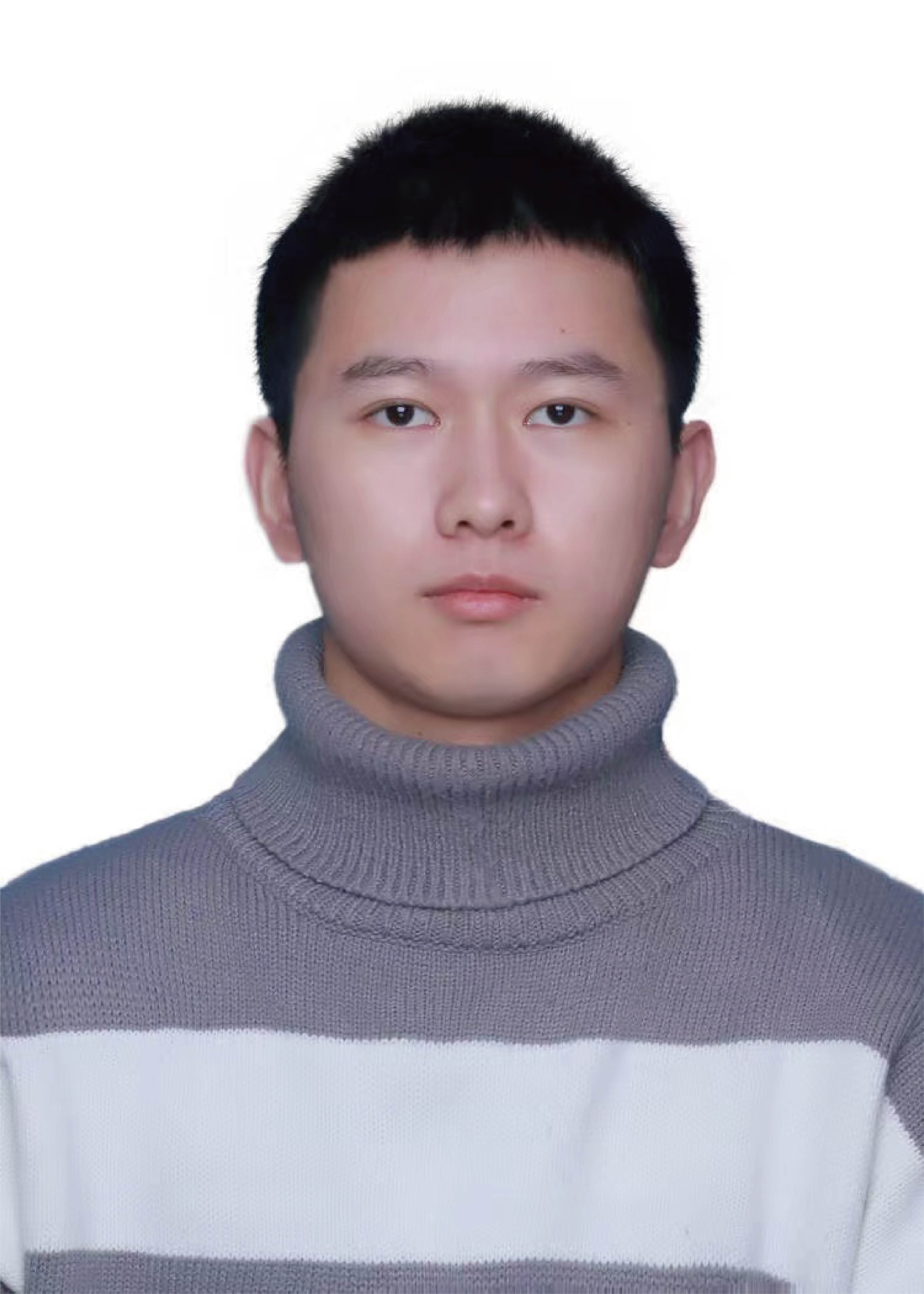}}]{Jie Wang}
received a BS degree from Hunan University of Chinese Medicine. He is currently working toward the MS degree in Electronic Information at the School of Automation and Artificial Intelligence, Nanjing University of Posts and Telecommunications. His research interests include 3D computer vision and digital humans.
\end{IEEEbiography}

\begin{IEEEbiography}[{\includegraphics[width=1in,height=1.25in,clip,keepaspectratio]{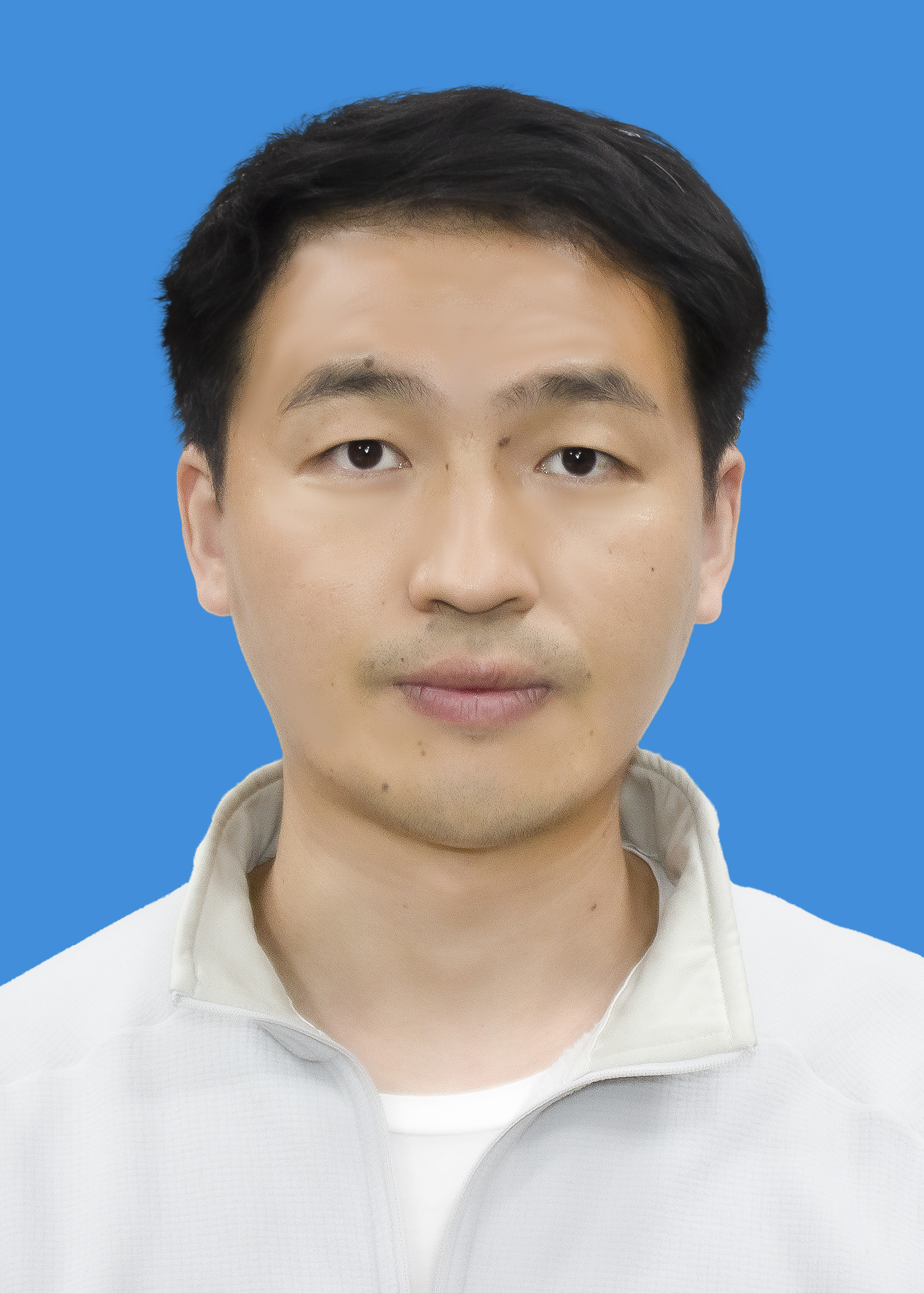}}]{Jiu-Cheng Xie} received his PhD degree in Computer and Information Science from the University of Macau in 2022, with a year of study at the Hong Kong Polytechnic University through a doctoral joint training program. He currently holds a lecturer position at the School of Automation and Artificial Intelligence, Nanjing University of Posts and Telecommunications, China. His research interests center on human-centric topics, including analysis, 3D reconstruction, animation, and their interactions with the environment.
\end{IEEEbiography}

\begin{IEEEbiography}[{\includegraphics[width=1in,height=1.25in,clip,keepaspectratio]{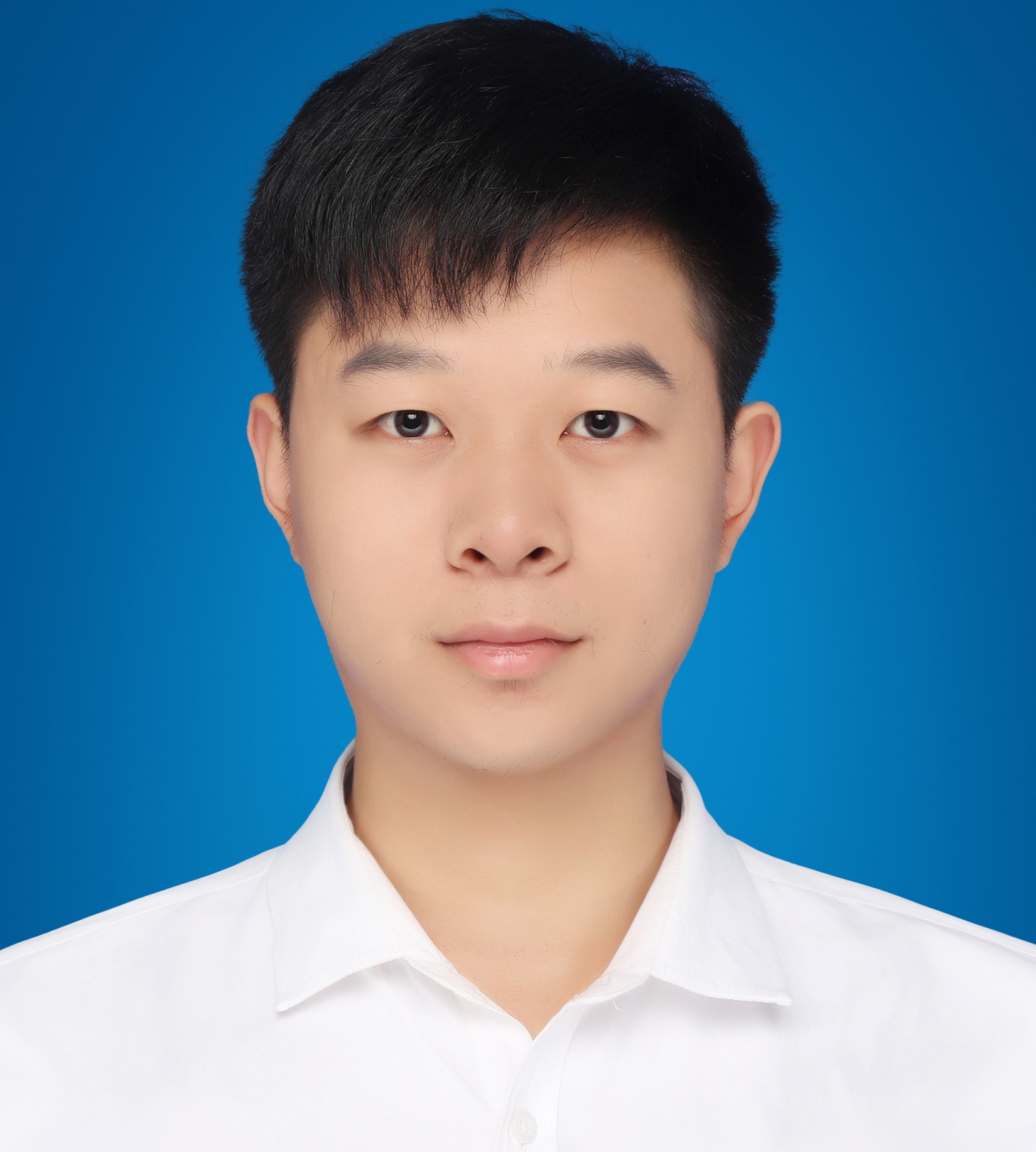}}]{Xianyan Li}
received a BS degree from Yantai University. He is currently working toward the MS degree in Electronic Information at the School of Automation and Artificial Intelligence, Nanjing University of Posts and Telecommunications. His research interests include virtual reality and intelligent robotics.
\end{IEEEbiography}

\begin{IEEEbiography}[{\includegraphics[width=1in,height=1.25in,clip,keepaspectratio]{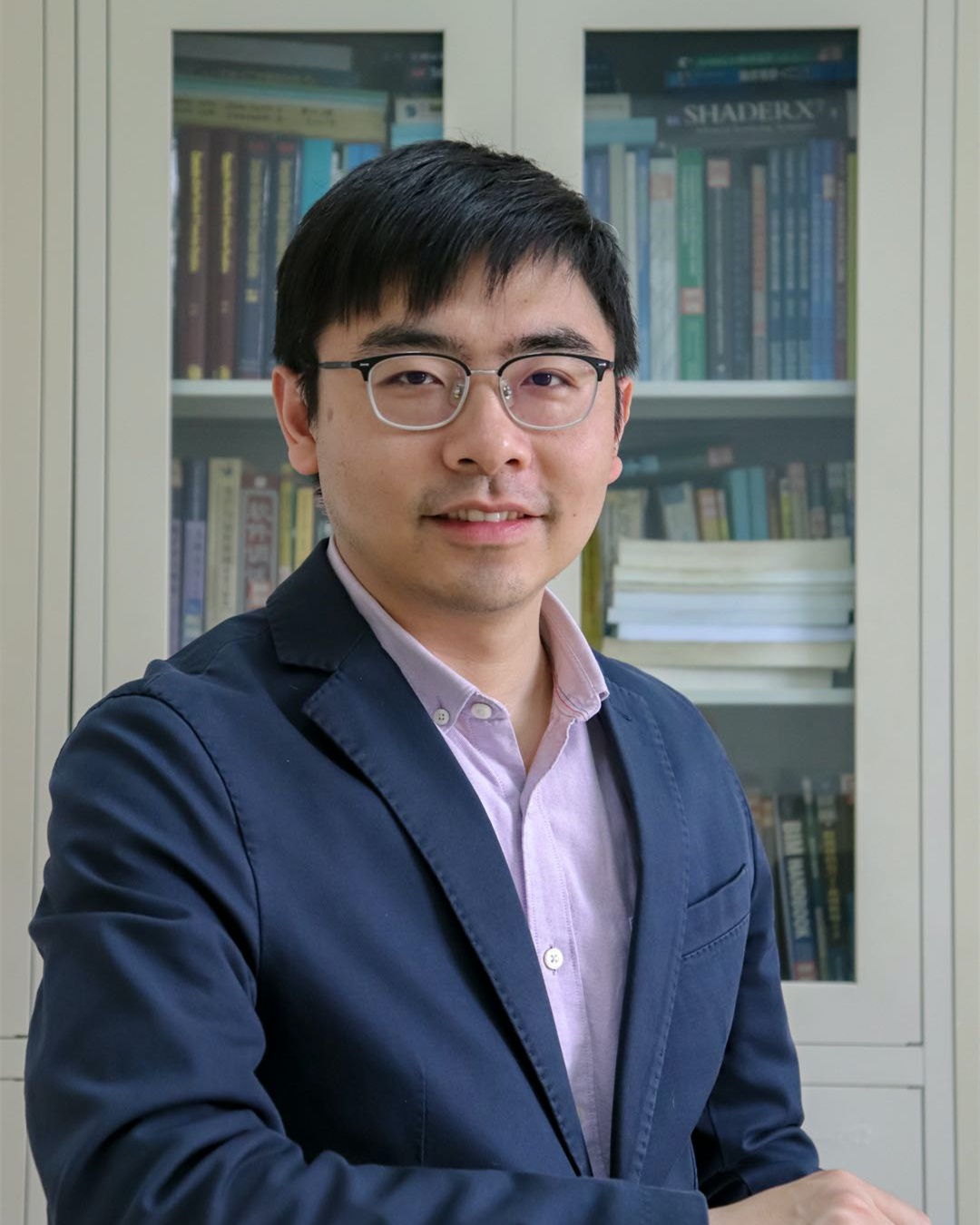}}]{Feng Xu} received the BS degree in physics from Tsinghua University, Beijing, China, in 2007, and the PhD degree in automation from Tsinghua University, Beijing, China, in 2012. He is currently an associate professor with the School of Software, Tsinghua University, China. His research interests include facial animation, performance capture, and 3D reconstruction.
\end{IEEEbiography}

\begin{IEEEbiography}[{\includegraphics[width=1in,height=1.25in,clip,keepaspectratio]{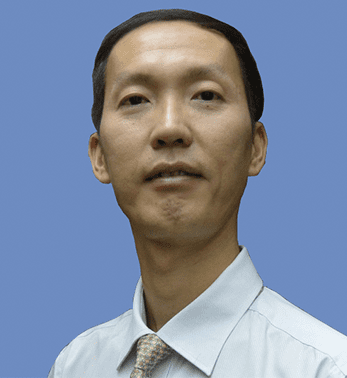}}]{Chi-Man Pun} received the B.Sc. and M.Sc. degrees in software engineering from the University of Macau in 1995 and 1998, respectively, and the Ph.D. degree in computer science and engineering from The Chinese University of Hong Kong in 2002. He was the Head of the Department of Computer and Information Science from 2014 to 2019. He is currently a Professor of computer and information science and in charge of the Image Processing and Pattern Recognition Laboratory, Faculty of Science and Technology, University of Macau. He has investigated many externally funded research projects as a PI, and has authored/coauthored more than 200 refereed papers in many top-tier journals and conferences. His research interests include image processing and pattern recognition; multimedia information security, forensic and privacy; adversarial machine learning and AI security. 
\end{IEEEbiography}

\begin{IEEEbiography}[{\includegraphics[width=1in,height=1.25in,clip,keepaspectratio]{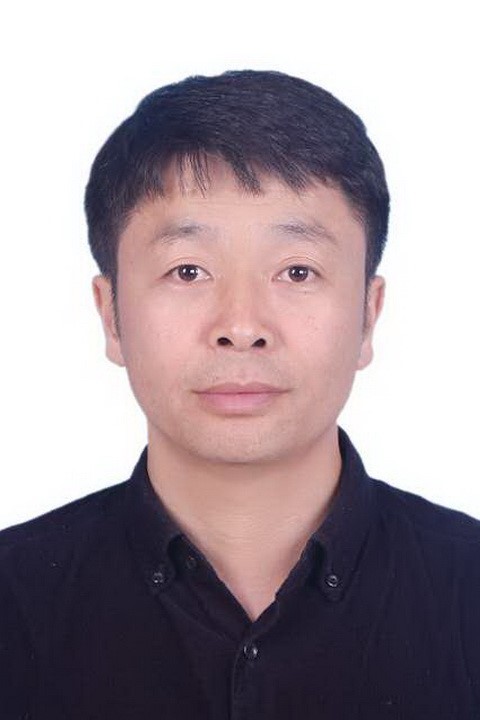}}]{Hao Gao} is currently a professor with the School of Automation, School of Artificial Intelligence, Nanjing University of Posts and Telecommunications, Nanjing, China. His research interests include artificial intelligence and computer vision, and he has authored or co-authored more than 50 international journal and conference papers. Prof. Gao was the Editorial Member and referee for many international journals.
\end{IEEEbiography}

\newpage

\vfill

\end{document}

\title{Supplementary Material - GaussianHead: High-fidelity Head Avatars with Learnable Gaussian Derivation}

\author{Jie Wang, Jiu-Cheng Xie, Xianyan Li, Feng Xu, Chi-Man Pun,   and Hao Gao% <-this % stops a space
\IEEEcompsocitemizethanks{\IEEEcompsocthanksitem 
Jie Wang, Jiu-Cheng Xie Xianyan Li and Hao Gao are with the School of Automation and the School of Artificial Intelligence, Nanjing University of Posts and Telecommunications, Nanjing, 210023, China.
E-mail: chieh.wangs@gmail.com,
jiuchengxie@gmail.com, 974598lxy@gmail.com,
tsgaohao@gmail.com.

\IEEEcompsocthanksitem Feng Xu is with the School of Software and BNRist, Tsinghua University, Beijing 100084, China.
E-mail: xufeng2003@gmail.com.

\IEEEcompsocthanksitem Chi-Man Pun is with the Department of Computer and Information Science, University of Macau, Taipa, Macau.  
E-mail: cmpun@um.edu.mo.

% \IEEEcompsocthanksitem xxx is xxx. E-mail: xxx@dlut.edu.cn.

\IEEEcompsocthanksitem Jiu-cheng Xie and Hao Gao are the corresponding authors.}}

\markboth{IEEE Trans Vis Comput Graph,~Vol.~X, No.~X, XXX}%
{Shell \MakeLowercase{\textit{et al.}}: Bare Demo of IEEEtran.cls for Computer Society Journals}
\maketitle
\IEEEdisplaynontitleabstractindextext

% \fi
\IEEEpeerreviewmaketitle

% \IEEEraisesectionheading{\section{NETWORK ARCHITECTURE}\label{sec:network}}
\section{NETWORK ARCHITECTURE}\label{sec:network}
\begin{figure}[tb] 
\centering
    \includegraphics[width=0.5\textwidth]{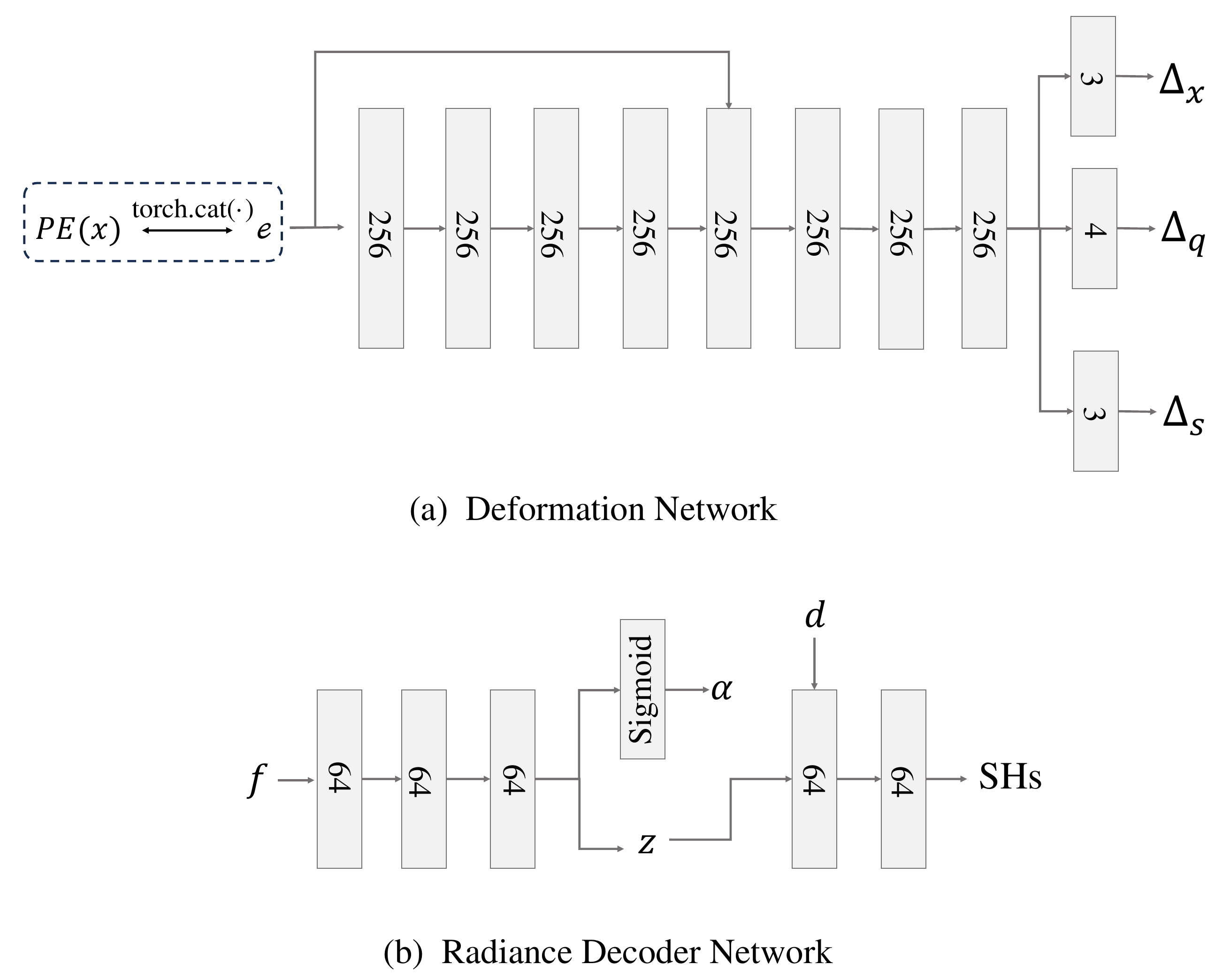}
    \caption{Diagrams of our network architecture. $PE(\cdot)$ and $e$ represent position encoding and expression parameters, respectively.} 
    \label{fig:network}
\end{figure}

Details of the network architecture are shown in Fig. \ref{fig:network}. The main parameters to be optimized in our framework include the mean position $\bm{x}$, quaternion-formed rotation $\bm{q}$, and scale $\bm{s}$ of initial 3D Gaussians, as well as a motion deformation MLP, a single-resolution tri-plane feature container, a radiance decoder MLP, and a set of learnable unit quaternions $\bm{r}$ used for deriving each core Gaussian in canonical space. 

The deformation network has two inputs: the latent code of $\bm{x}$ processed by imposing the positional encoding with a frequency of 10 on it and the tracked expression parameters $\bm{e}$. This module consists of 8 linear layers with a unified width of 256 and skip connections added to the fourth layer. The final linear layer is connected to three parallel heads, each of which is activated using ReLU. They respectively output offsets of the position ${\Delta}{\bm{x}}$, rotation ${\Delta}{\bm{q}}$, and scale ${\Delta}_{\bm{s}}$ relative to their initial values. 

% Positional encoding with a frequency of 10 is used for the initial Gaussians' mean position $\bm{x}$ in the input deformation network. The deformation network consists of 8 linear layers with a width of 256, and skip connections are added in the fourth layer. After the eighth linear layer, three output heads, respectively, output the position offsets ${\Delta}{\bm{x}}$, rotation offsets ${\Delta}{\bm{q}}$, and scale offsets ${\Delta}_{\bm{s}}$ of the initial 3D Gaussians, all of which are activated using ReLU. The learning rate for the initial 3D Gaussian's position is set to $1.6 \times 10^{-4}$, linearly decaying to $1.6 \times 10^{-6}$ over training steps; rotation and scale have a learning rate of $1 \times 10^{-3}$. The initial learning rate for the motion deformation network is set to $8 \times 10^{-4}$, and it gradually decays to $8 \times 10^{-6}$ as the training progressed.

The parameters of the single-resolution tri-plane are initialized with a uniform distribution in the range [0.1, 0.5]. The resolution is set to $64\times64$. For the canonical features $\bm{f}$, an opacity network consisting of 3 linear layers with a width of 64 is used to process them. The opacity $\alpha$ is activated by the sigmoid function. The intermediate variables $\bm{z}$ are concatenated with the view direction $\bm{d}$ and fed into a color network with a depth of 2 and a width of 64. The ReLU function is used for activation purposes, yielding spherical harmonic coefficients $Y_{lm}$.

%\vfill

